\documentclass[a4paper,twoside]{article}

\usepackage{epsfig}
\usepackage{calc}
\usepackage{amssymb}
\usepackage{amstext}
\usepackage{amsmath}
\usepackage{amsthm}
\usepackage{multicol}
\usepackage{pslatex}
\usepackage{apalike}
\usepackage{algorithm2e}
\usepackage[bottom]{footmisc}

\usepackage{comment}
\usepackage{xcolor}
\usepackage{amsfonts}
\usepackage{subfig}
\usepackage{tikz}
\usetikzlibrary{arrows}
\usetikzlibrary{arrows.meta}

\usepackage{SCITEPRESS}     

\usepackage[pagebackref=true,breaklinks=true,colorlinks,bookmarks=false]{hyperref}

\DeclareMathOperator*{\argmax}{arg\, max}
\DeclareMathOperator*{\argmin}{arg\, min}

\newcommand{\ada}{\text{ADA}^*}
\newcommand{\auc}{\text{AUROC}}
\newcommand{\tpr}{\text{TPR}_{5\%}}

\definecolor{isabelline}{rgb}{0.96, 0.94, 0.93}
\definecolor{aliceblue}{rgb}{0.94, 0.97, 1.0}
\definecolor{lavenderblush}{rgb}{1.0, 0.94, 0.96}
\definecolor{magnolia}{rgb}{0.97, 0.96, 1.0}

\begin{document}

\title{Uncertainty-Based Detection of Adversarial Attacks in~Semantic~Segmentation}

\author{\authorname{Kira Maag\sup{1} and Asja Fischer\sup{2} }
\affiliation{\sup{1}Technical University of Berlin, Germany}
\affiliation{\sup{2}Ruhr University Bochum, Germany}
\email{maag@tu-berlin.de, asja.fischer@rub.de}
}

\keywords{Deep Learning, Semantic Segmentation, Adversarial Attacks, Detection.}

\abstract{State-of-the-art deep neural networks have proven to be highly powerful in a broad range of tasks, including semantic image segmentation. However, these networks are vulnerable against adversarial attacks, i.e., non-perceptible perturbations added to the input image causing incorrect predictions, which is hazardous in safety-critical applications like automated driving. Adversarial examples and defense strategies are well studied for the image classification task, while there has been limited research in the context of semantic segmentation. First works however show that the segmentation outcome can be severely distorted by adversarial attacks. In this work, we introduce an uncertainty-based approach for the detection of adversarial attacks in semantic segmentation. We observe that uncertainty as for example captured by the entropy of the output distribution behaves differently on clean and perturbed images and leverage this property to distinguish between the two cases. Our method works in a light-weight and post-processing manner, i.e., we do not modify the model or need knowledge of the process used for generating adversarial examples. In a thorough empirical analysis, we demonstrate the ability of our approach to detect perturbed images across multiple types of adversarial attacks.}

\onecolumn \maketitle \normalsize \setcounter{footnote}{0} \vfill

\section{\uppercase{Introduction}}
\begin{figure}
    \centering
    \subfloat[][Input image (perturbed half on right hand side)]{\includegraphics[trim=0 824 0 80,clip,width=0.46\textwidth]{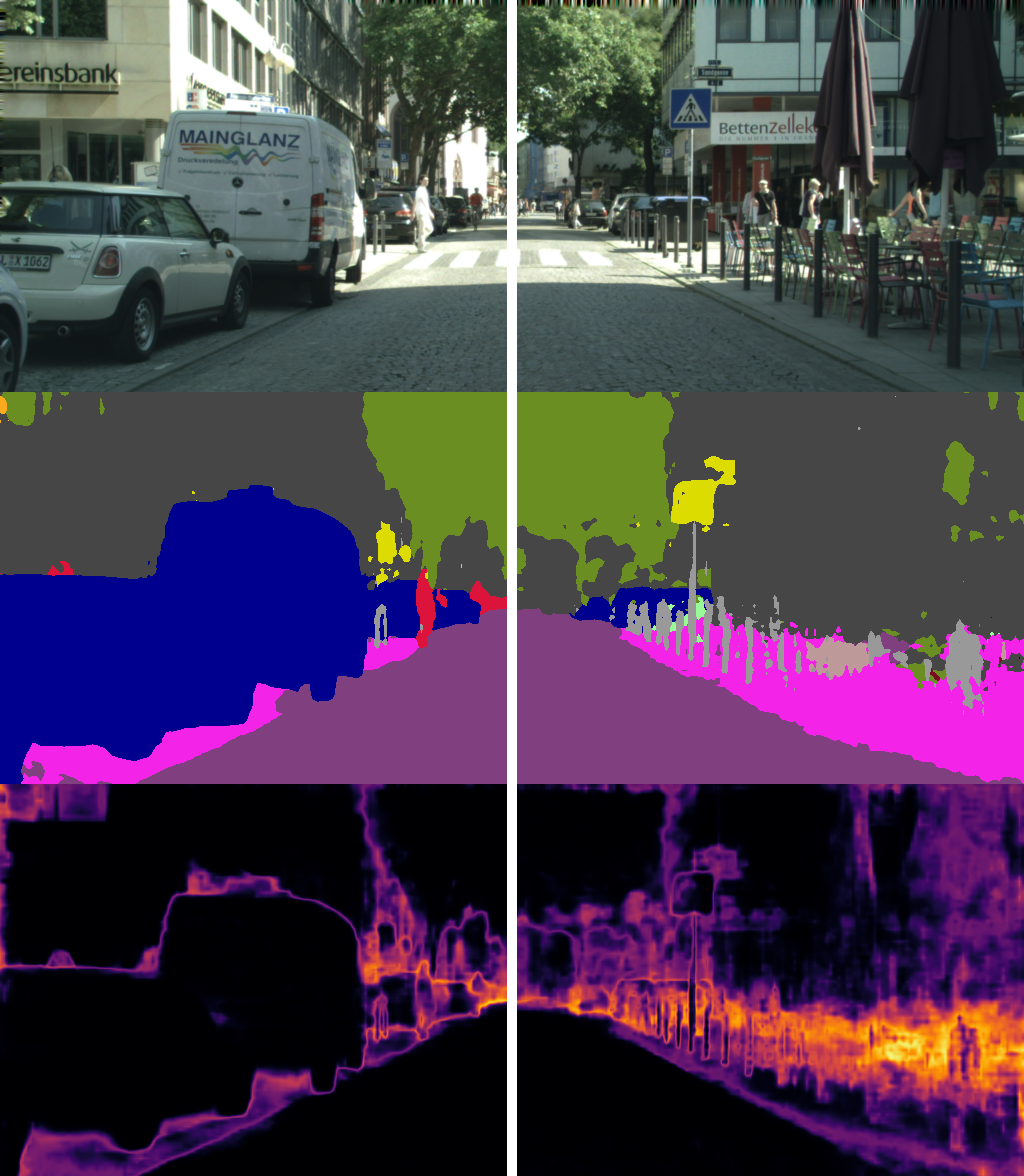}} \\
    \vspace{-2ex}
    \subfloat[][Semantic Segmentation prediction]{\includegraphics[trim=0 432 0 472,clip,width=0.46\textwidth]{figs/pred_heat.png}} \\
    \vspace{-2ex}
    \subfloat[][Entropy heatmap]{\includegraphics[trim=0 40 0 864,clip,width=0.46\textwidth]{figs/pred_heat.png}}
    \caption{Semantic segmentation prediction and entropy heatmap for clean (\emph{left}) and perturbed image (\emph{right}) generated by a dynamic target attack for hiding pedestrians.}
    \label{fig:pred_heat}
\end{figure}

In recent years, deep neural networks (DNNs) have demonstrated outstanding performance and have proven to be highly expressive in a broad range of tasks, including semantic image segmentation \cite{Chen2018,Pan2022}. Semantic segmentation aims at segmenting objects in an image by assigning each pixel to a fixed and predefined set of semantic classes, providing comprehensive and precise information about the given scene. However, DNNs are vulnerable to \emph{adversarial attacks} \cite{Bar2021} which is very hazardous in safety-related applications like automated driving. Adversarial attacks are small perturbations added to the input image causing the DNN to perform incorrect predictions at test time. 
The perturbations are not perceptible to humans, making the detection of these examples very challenging, see for example Figure~\ref{fig:pred_heat}.
This undesirable property of DNNs is a major security concern in real world applications.  Hence, developing efficient strategies against adversarial attacks is of high importance. Such strategies can either increase the robustness of DNNs making it more difficult to generate adversarial examples (defense) or build on approaches to detect adversarial attacks (detection).

Adversarial attacks have attracted much attention, and numerous attacks as well as detection strategies have been proposed \cite{Khamaiseh2022}. However, adversarial examples have not been analyzed extensively beyond standard image classification models, often using small datasets such as MNIST \cite{LeCun2010} or CIFAR10 \cite{Krizhevsky2009}. The vulnerability of modern DNNs to attacks in more complex tasks like semantic segmentation in the context of real datasets from different domains has been rather poorly explored. 
The attacks which have been tested so far on semantic segmentation networks can be divided roughly into three categories. 
The first approaches transfer common attacks from image classification to semantic segmentation, i.e., pixel-wise classification \cite{Agnihotri2023,Gu2022,Rony2022}. 
Other works \cite{Cisse2017,Xie2017} have presented attacks specifically designed for the task of semantic segmentation, either attacking the input in a way that leads to the prediction of some predefined and image-unrelated segmentation mask or the complete deletion of one segmentation class \cite{Metzen2017}.
Such attacks are more difficult to detect than attacks that perturb each pixel independently. 
The third group of attacks creates rectangular patches smaller than the input size, which leads to an erroneous prediction of the entire image \cite{Nakka2020,Nesti2022}. 
Defense methods aim to be robust against such attacks, i.e., to achieve high prediction accuracy even on perturbed images. For semantic segmentation tasks, defense approaches are often robust against only one type of attack \cite{Arnab2018,Klingner2020,Yatsura2022}. 
In contrast, detection methods aim at classifying an input as clean or perturbed based on the output of the segmentation model \cite{Xiao2018}.

In this paper, we present an uncertainty-based approach for detecting several kinds of adversarial attacks on semantic image segmentation models. Uncertainty information has already been exploited for the detection of adversarial attacks on classification DNNs, but has not been investigated in the context of segmentation so far. 
In \cite{Feinman2017} an approximation to Bayesian inference (Monte-Carlo Dropout) is proposed, which is widely employed to estimate model uncertainty, for the detection of adversarial attacks. 
The gradient-based approach introduced in \cite{Michel2022} generates salient features which are used to train a detector.
While both methods require access to the inside of the 
model, our approach can be applied as a post-processing step using only information of the network output. We construct features per image based on uncertainty information provided by the DNN such as the entropy of the output distribution. In Figure~\ref{fig:pred_heat} (c), the entropy heatmaps for a clean (left) and a perturbed image (right) are shown, indicating high uncertainties in the attacked regions, motivating the use of uncertainty information to separate clean and perturbed images. 
On the one hand, these features for clean and perturbed inputs are fed into an one-class support vector machine that performs unsupervised novelty detection \cite{Weerasinghe2018}. 
On the other hand, we train a logistic regression model with clean and perturbed data for classification.
The perturbed data used during training is generated by only one kind of adversarial attack method, while the detector is applied to identify adversarial examples of other methods.
Our approach neither modifies the semantic segmentation model nor requires knowledge of the process for generating adversarial examples. We only assume that our post-processing model is kept private while the attacker may have full access to the semantic segmentation model. 

In our tests, we employ state-of-the-art semantic segmentation networks \cite{Chen2018,Pan2022} applied to the Cityscapes \cite{Cordts2016} as well as Pascal VOC2012 dataset \cite{Everingham2012} demonstrating our adversarial attack detection performance. To this end, we consider different types of attackers, from pixel-level attacks designed for image classification \cite{Goodfellow2015,Kurakin2017} to pixel-wise attacks developed for semantic segmentation \cite{Metzen2017} and patch-based ones \cite{Nesti2022}.
The source code of our method is publicly available at \url{https://github.com/kmaag/Adversarial-Attack-Detection-Uncertainty}.
Our contributions are summarized as follows:
\begin{itemize}
    \item We introduce a new uncertainty-based approach for the detection of adversarial attacks for the semantic image segmentation task. In a thorough empirical analysis, we demonstrate the capability of uncertainty measures to distinguish between clean and perturbed images. 
    Our approach serves as a light-weight post-processing step, i.e., we do not modify the model or need knowledge of the process for generating adversarial examples.
    \item For the first time, we present a detection method that was not designed for a specific adversarial attack, rather has a high detection capability across multiple types. We achieve averaged detection accuracy values of up to $100\%$ for different network architectures and datasets.
\end{itemize}
%
%
%
\section{\uppercase{Related Work}}\label{sec:rel_work}
In this section, we discuss the related works on defense and detection methods for the semantic segmentation task. Defense methods aim to achieve high prediction accuracy even on perturbed images, while detection methods classify the model input as clean or attacked image. 
A dynamic divide-and-conquer strategy \cite{Xu2021} and multi-task training \cite{Klingner2020}, which extends supervised semantic segmentation by a self-supervised monocular depth estimation using unlabeled videos, are considered as adversarial training approaches enhancing the robustness of the networks.
Another kind of defense strategy is input denoising to remove the perturbation from the input without the necessity to re-train the model. In \cite{Bar2021} image quilting and the non-local means algorithm are presented as input transformation techniques. To denoise the perturbation and restore the original image, a denoise autoencoder is used in \cite{Cho2020}. The demasked smoothing technique, introduced in \cite{Yatsura2022}, reconstructs masked regions of each image based on the available information with an inpainting model defending against patch attacks.
Another possibility to increase the robustness of the model is during inference. In \cite{Arnab2018} is shown how mean-field inference and multi-scale processing naturally form an adversarial defense. The non-local context encoder proposed in \cite{He2019} models spatial dependencies and encodes global contexts for strengthening feature activations. From all pyramid features multi-scale information is fused to refine the prediction and create segmentation.
The presented works up to now are defense methods improving the robustness of the model.
To the best of our knowledge so far only one work focuses on detecting adversarial attacks on segmentation models, i.e., the patch-wise spatial consistency check which is introduced in \cite{Xiao2018}.

The described defense approaches are created for and tested only on a specific type of attack. The problem is that you assume a high model robustness, however, the defense method may perform poorly on new unseen attacks and does not provide a statement about this insecurity. Therefore, we present an uncertainty-based detection approach which shows strong results over several types of adversarial attacks.
The presented detection approach \cite{Xiao2018} is only tested on perturbed images attacked in such a way that a selected image is predicted. The spatial consistency check randomly selects overlapping patches to obtain pixel-wise confidence vectors. In contrast, we use only information of the network output from one inference and not from (computationally
expensive) multiple runs of the network. For these reasons, the detection approach introduced in \cite{Xiao2018} cannot be considered as a suitable baseline.

Post-processing classification models as well as simple output-based methods are used for false positive detection \cite{Maag2019,Maag2023} and out-of-distribution segmentation \cite{Maag2022,Hendrycks2016}, but have not been investigated
for adversarial examples.
%
%
%
\section{\uppercase{Adversarial Attacks}}\label{sec:attacks}
For the generation of adversarial examples, we distinguish between \emph{white} and \emph{black box} attacks. White box attacks are created based on information of the victim model, i.e., the adversarial attacker has access to the full model, including its parameters, and knows the loss function used for training.
In contrast, black box attackers have zero knowledge about the victim model. The idea behind these type of attacks is transferability, i.e., an adversarial example generated from another model works well with the victim one.
The attacks described in the following belong to the white box setting
and were proposed to attack semantic segmentation models. 
%
%
\paragraph{Attacks on Pixel-wise Classification}
The attacks described in this paragraph were originally developed for image classification and were (in a modified version) applied to semantic segmentation. 
For semantic segmentation, given an image $x$ a neural network with parameters $w$ provides pixel-wise probability distributions $f(x;w)_{ij}$ over a label space $C = \{y_{1}, \ldots, y_{c} \}$ per spatial dimension $(i,j)$. 
The single-step untargeted \emph{fast gradient sign method} (FGSM, \cite{Goodfellow2015}) creates adversarial examples by adding perturbations to the pixels of an image $x$ with (one-hot encoded) label $y$ that leads to an increase of the loss $L$ (here cross entropy), that is
\begin{equation}
    x_{ij}^{\mathit{adv}} = x_{ij} + \varepsilon \cdot \text{sign}(\nabla_x L_{ij}(f(x;w)_{ij},y_{ij})) \enspace, 
\end{equation}
where $\varepsilon$ is the magnitude of perturbation. 
The single-step targeted attack with target label $y_{ll}$ instead decreases the loss for the target label and is given by \begin{equation}\label{eq:fgsm_ll}
    x_{ij}^{\mathit{adv}} = x_{ij} - \varepsilon \cdot \text{sign}(\nabla_x L_{ij}(f(x;w)_{ij},y_{ij}^{ll})) \enspace . 
\end{equation}
Following the convention, the least likely class predicted by the model is chosen as target class.
This attack is extended in \cite{Kurakin2017} in an iterative manner (I-FGSM) to increase the perturbation strength
\begin{align}
    x^{\mathit{adv}}_{ij,t+1} = & \\ \text{clip}_{x,\varepsilon} \big(x^{\mathit{adv}}_{ij,t}   + \alpha \cdot \text{sign}(\nabla_{x^{\mathit{adv}}_{t}} \notag
    & \ L_{ij}(f(x^{\mathit{adv}}_{t};w)_{ij},y_{ij})) \big) 
\end{align}
with $x^{\mathit{adv}}_{0} = x$, step size $\alpha$, and a clip function ensuring that $x^{\mathit{adv}}_{t} \in [x-\varepsilon, x+\varepsilon]$. The targeted case (see eq.~(\ref{eq:fgsm_ll})) can be formulated analogously. 
Based on these attacks, further methods for pixel-wise perturbations in the classification context have been proposed such as projected gradient descent \cite{Madry2018,Bryniarski2022} and DeepFool \cite{Moosavi2016}. Some of these approaches have been further developed and adapted to semantic segmentation \cite{Agnihotri2023,Gu2022,Rony2022}.
%
%
\paragraph{Stationary Segmentation Mask Attacks}
Another type of attacks are so called \emph{stationary segmentation mask methods} (SSMM) where the pixels of a whole image are iteratively attacked until most of the pixels have been mis-classified into the target class \cite{Cisse2017,Xie2017}. 
For each spatial dimension $(i,j) \in \mathcal{I}$, the loss function per image $x$ is given by
\begin{equation}\label{eq:loss}
    L(f(x;w),y) = \frac{1}{|\mathcal{I}|} \sum_{(i,j) \in \mathcal{I}} L_{ij}(f(x;w)_{ij},y_{ij}) \, . 
\end{equation}
In \cite{Metzen2017}, the universal perturbation is introduced to achieve real-time performance for the attack at test time. To this end, training inputs $D^{\text{train}}= \{ x^{(k)}, y^{(k),\text{target}} \}_{k=1}^m $ are generated where $y^{(k),\text{target}}$ defines a fixed target segmentation. 
The universal noise in iteration $t+1$ is computed by 
\begin{align}
    \xi_{t+1} = & \ \text{clip}_{\varepsilon} \big(\xi_{t}  \\- \alpha \cdot \text{sign}( \frac{1}{m} \sum_{k=1}^m \nabla_x \notag 
    & \ L(f(x^{(k)} + \xi_{t};w),y^{(k),\text{target}}) \big)  
\end{align}
with $\xi_{0} = 0$. The loss of pixels which are predicted as belonging to the desired target class with a confidence above a threshold $\tau$ are set to $0$.
At test time, this noise is added to the input image and does not require multiple calculations of the backward pass. 

The \emph{dynamic nearest neighbor method} (DNNM) presented in \cite{Metzen2017}  aims to keep the network’s segmentation unchanged but to remove a desired target class. Let $o$ be the object class being deleted and $\hat y(x)_{ij} =\argmax_{y\in\mathcal{C}} f(x;w)_{ij}^{y} $ the predicted class, where $f(x;w)_{ij}^{y}$ denotes the probability the model assigns for the pixel at position $(i,j)$ to belong to class $y$, then $\mathcal{I}_o = \{ (i,j)| \hat y(x)_{ij} = o \}$ and $\mathcal{I}_{\bar{o}} = \mathcal{I} \setminus \mathcal{I}_o$. The target label is chosen by $y_{ij}^{\text{target}} = \hat y(x)_{i'j'}$ with $\argmin_{(i',j') \in \mathcal{I}_{\bar{o}}} (i'-i)^2+(j'-j)^2$ for all $(i,j) \in \mathcal{I}_o$ and $y_{ij}^{\text{target}} = \hat y(x)_{ij}$ for all $(i,j) \in \mathcal{I}_{\bar{o}}$. 
Since the loss function described in eq.~(\ref{eq:loss}) weights all pixels equally though both objectives, i.e., hiding a object class and being unobtrusive are not necessarily equally important, a modified version of the loss function with weighting parameter $\omega$ is given by
\begin{align}
    L^{\omega} (f(x;w),y) & = \frac{1}{|\mathcal{I}|} ( \omega \sum_{(i,j) \in \mathcal{I}_o} L_{ij} (f(x;w)_{ij},y_{ij}^{\text{target}}) \notag \\
    & + (1-\omega) \sum_{(i,j) \in \mathcal{I}_{\bar{o}}} L_{ij} (f(x;w)_{ij},y_{ij}^{\text{target}}) ) \, .
\end{align}
Note, the universal perturbation can also be computed for the DNNM.
%
%
\paragraph{Patch-based Attacks}
The idea behind \emph{patch attacks} is that perturbing a small region of the image causes prediction errors in a much larger region \cite{Nakka2020}. In \cite{Nesti2022}, the \emph{expectation over transformation} (EOT)-based patch attack is introduced to create robust adversarial examples, i.e., individual adversarial examples that are at the same time adversarial over a range of transformations. Transformations occurring in the real world are for instance angle and viewpoint changes. These perturbations are modeled within the optimization procedure and an extension of the pixel-wise cross entropy loss is additionally presented in \cite{Nesti2022} to enable crafting strong patches for the semantic segmentation setting. 
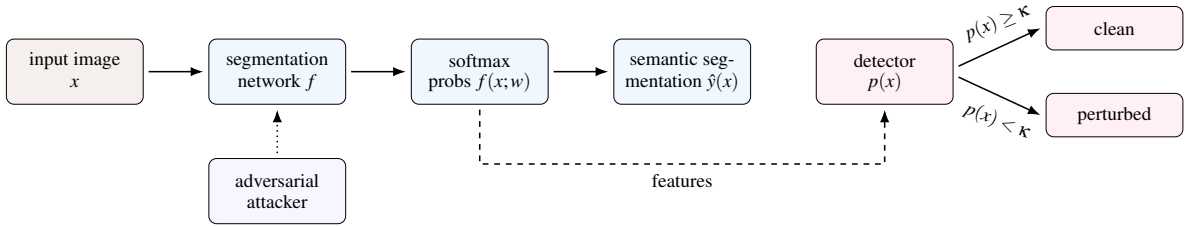
\begin{figure*}
    \centering
    \scalebox{0.72}{\begin{tikzpicture}

\draw[fill=isabelline,rounded corners] (0, 0) rectangle (2.5,1.2) {};
\node at (1.25,0.8) {input image};
\node at (1.25,0.4) {$x$};

\draw [-Latex,thick] (2.6,0.6) -- (3.6,0.6);

\draw[fill=aliceblue,rounded corners] (3.7, 0) rectangle (6.2,1.2) {};
\node at (4.95,0.8) {segmentation};
\node at (4.95,0.4) {network $f$};

\draw [-Latex,thick] (6.3,0.6) -- (7.3,0.6);

\draw[fill=aliceblue,rounded corners] (7.4, 0) rectangle (9.9,1.2) {};
\node at (8.65,0.8) {softmax};
\node at (8.65,0.4) {probs $f(x;w)$};

\draw [-Latex,thick] (10.0,0.6) -- (11.0,0.6);

\draw[fill=aliceblue,rounded corners] (11.1, 0) rectangle (13.6,1.2) {};
\node at (12.35,0.8) {semantic seg-};
\node at (12.35,0.4) {mentation $\hat{y}(x)$};


\draw[fill=lavenderblush,rounded corners] (14.8, 0) rectangle (17.3,1.2) {};
\node at (16.05,0.8) {detector};
\node at (16.05,0.4) {$p(x)$};

\draw [-Latex,thick] (17.4,0.7) -- (18.9,1.4);
\draw [-Latex,thick] (17.4,0.5) -- (18.9,-0.2);

\node[rotate=22] at (18.1,1.5) {$p(x) \geq \kappa$};
\node[rotate=-22] at (18.1,-0.3) {$p(x) < \kappa$};

\draw[fill=lavenderblush,rounded corners] (19.0, 1) rectangle (21.5,1.8) {};
\node at (20.25,1.4) {clean};

\draw[fill=lavenderblush,rounded corners] (19.0, 0.2) rectangle (21.5,-0.6) {};
\node at (20.25,-0.2) {perturbed};

\draw [-Latex,thick,dotted] (4.95,-0.9) -- (4.95,-0.1);

\draw[fill=magnolia,rounded corners] (3.7, -2.2) rectangle (6.2,-1.0) {};
\node at (4.95,-1.4) {adversarial};
\node at (4.95,-1.8) {attacker};

\draw [-Latex,thick, dashed] (8.65,-0.1) -- (8.65,-1.1) -- (16.05,-1.1) -- (16.05,-0.1);
\node at (12.35,-1.4) {features};

\end{tikzpicture}}
    \caption{Schematic illustration of our detection method. The adversarial attacker can have full access to the semantic segmentation model. Information from the network output is extracted to construct the features which serve as input to the detector model classifying between clean and perturbed images.}
    \label{fig:method}
\end{figure*}
%
%
%
\section{\uppercase{Detection Method}}\label{sec:method} 
Our method does not alter the semantic segmentation model, nor does it require knowledge of the adversarial example generation process. While the attacker may have full access to the semantic segmentation model, we only assume that our post-processing model is kept secret or not attacked. Our approach can be applied to any semantic segmentation network serving as a post-processing step using only information of the network output. In Figure~\ref{fig:method} an overview of our approach is given. 

The degree of uncertainty in a semantic segmentation prediction is quantified by pixel-wise dispersion measures like the entropy
\begin{equation}
    E(x)_{ij} =-\sum_{y\in \mathcal{C}}f(x;w)_{ij}^{y}\log f(x;w)_{ij}^{y} \enspace ,
\end{equation}
the variation ratio 
\begin{equation}
    V(x)_{ij} = 1 - \max_{y\in\mathcal{C}} f(x;w)_{ij}^{y}\enspace,
\end{equation}
or the probability margin 
\begin{equation}
    M(x)_{ij} = V(x)_{ij} + \max_{y\in\mathcal{C}\setminus\{\hat y(x)_{ij}\}} f(x;w)_{ij}^{y} \enspace .
\end{equation}
The entropy heatmaps for a clean (left) and a perturbed image (right) are shown in Figure~\ref{fig:pred_heat} (c) indicating that higher uncertainties occur in the attacked regions which motivates the use of uncertainty information to separate clean and perturbed data.
To obtain uncertainty features per image from these pixel-wise dispersion measures, we aggregate them over a whole image by calculating the averages $\bar D = 1 / |\mathcal{I}|\sum_{(i,j) \in \mathcal{I}} D(x)_{ij}$
where $D \in \{E,V,M\}$.
Moreover, we obtain mean class probabilities for each class $y \in \{1,\ldots,C\}$
\begin{equation}
    P(y|x) = \frac{1}{|\mathcal{I}|} \sum_{(i,j) \in \mathcal{I}} f(x;w)_{ij}^{y} \, .
\end{equation}
The concatenation of this $|C|+3$ features forms the feature vectors used in the following. 
We compute these image-wise features for a set of benign (and adversarially changed) images, which are then used to train classification models providing per image a probability $p(x)$ of being clean (and not perturbed).
We classify $x$ as perturbed if $p(x) < \kappa$ and as clean if $p(x) \geq \kappa$, where $\kappa$ is a predefined detection threshold.
We explore different ways to construct such a classifier.

First, we consider two basic outlier detection techniques which only require benign data, i.e., an one-class support vector machine (OCSVM, \cite{Scholkopf1999}) and an approach for detecting outliers in a Gaussian distributed dataset learning an ellipse \cite{Rousseeuw1999}. 
Second, we consider the supervised logistic regression (LASSO, \cite{Tibshirani1996}) as classification model trained on the features extracted for clean and perturbed images. Importantly, we do not require knowledge of the adversarial example generation process used by the attacker, instead we use attacked data generated by any (other) adversarial attack (cross-attack). While the OCSVM and the ellipse approach are unsupervised and outlier detection is a difficult task, the supervised cross-attack method has the advantage of having already seen other types of perturbed data.
Third, we threshold only on the mean entropy $\bar{E}$ (which requires only to choose the threshold value) proposing a very basic uncertainty-based detector.
Note, applying our detection method is light-weight, i.e., the feature computation is inexpensive and classification models are trained in advance so that only one inference run is added after semantic segmentation inference.
\begin{figure*}
    \centering
    \subfloat[][Input image]{\includegraphics[trim=0 512 3072 0,clip,width=0.22\textwidth]{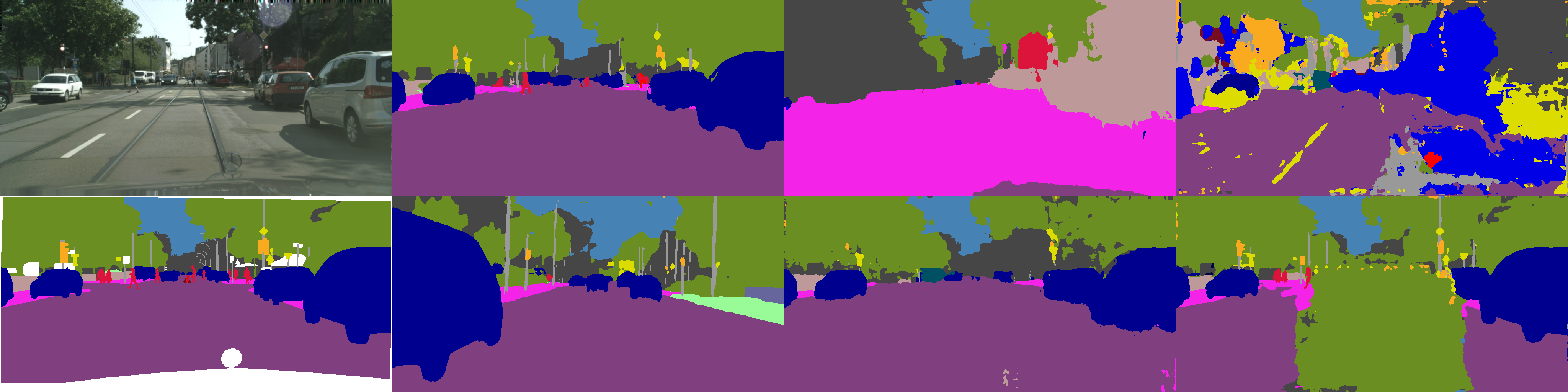}}
    \hspace{1ex}
    \subfloat[][Prediction]{\includegraphics[trim=1024 512 2048 0,clip,width=0.22\textwidth]{figs/preds_adv.png}}
    \hspace{1ex}
    \subfloat[][I-FGSM$_4$]{\includegraphics[trim=2048 512 1024 0,clip,width=0.22\textwidth]{figs/preds_adv.png}}
    \hspace{1ex}
    \subfloat[][I-FGSM$_4^{ll}$]{\includegraphics[trim=3072 512 0 0,clip,width=0.22\textwidth]{figs/preds_adv.png}} \\
    \vspace{-2ex}
    \subfloat[][Ground truth]{\includegraphics[trim=0 0 3072 512,clip,width=0.22\textwidth]{figs/preds_adv.png}}
    \hspace{1ex}
    \subfloat[][SSMM]{\includegraphics[trim=1024 0 2048 512,clip,width=0.22\textwidth]{figs/preds_adv.png}}
    \hspace{1ex}
    \subfloat[][DNNM]{\includegraphics[trim=2048 0 1024 512,clip,width=0.22\textwidth]{figs/preds_adv.png}}
    \hspace{1ex}
    \subfloat[][Patch attack]{\includegraphics[trim=3072 0 0 512,clip,width=0.22\textwidth]{figs/preds_adv.png}}
    \caption{Input image (a) with corresponding ground truth (e). Semantic segmentation prediction for clean (b) and perturbed image generated by an untargeted (c) and a targeted FGSM attack (d) as well as by SSMM (f), DNNM (g) and patch attack (h).}
    \label{fig:preds_adv}
\end{figure*}
%
%
%
\section{\uppercase{Experiments}}\label{sec:exp}
First, we present the experimental setting and then evaluate our adversarial detection approach.
%
%
\subsection{Experimental Setting}\label{sec:exp_setting}
%
%
\paragraph{Datasets}
We perform our tests on the Cityscapes \cite{Cordts2016} dataset for semantic segmentation in street and on the Pascal VOC2012 \cite{Everingham2012} (shorthand VOC) dataset of visual object classes in realistic scenes. 
The Cityscapes dataset consists of $2,\!975$ training and $500$ validation images of dense urban traffic in $18$ and $3$ different German towns, respectively.
The VOC dataset contains $1,\!464$ training and $1,\!449$ validation images with annotations for the various objects of categories person, animal, vehicle and indoor.
%
%
\paragraph{Segmentation Networks}
We consider the state-of-the-art DeepLabv3+ network \cite{Chen2018} with Xception65 backbone \cite{Chollet2017}. Trained on Cityscapes, we achieve a mean intersection over union (mIoU) value of $78.93$ on the validation set and trained on VOC, a validation mIoU value of $88.39$.
Moreover, we use the BiSeNet \cite{Yu2018} trained on Cityscapes obtaining a validation mIoU of $70.32$.
We consider also two real-time models for the Cityscapes dataset, the DDRNet \cite{Pan2022} achieving $77.8$ mIoU and the HRNet \cite{Wang2021} with $70.5$ mIoU.
%
%
\paragraph{Adversarial Attacks}
In many defense methods in semantic segmentation, the adapted FGSM and I-FGSM attack are employed \cite{Arnab2018,Bar2021,He2019,Klingner2020,Xu2021}. Thus, we study both attacks in our experiments with the parameter setting presented in \cite{Kurakin2017}. 
The magnitude of perturbation is given by $\varepsilon = \{ 4,8,16 \} $, the step size by $\alpha=1$ and the number of iterations is computed as $n = \min \{ \varepsilon+4, \lfloor 1.25\varepsilon \rfloor \} $. We denote the attack by FGSM$_{\varepsilon}^{\#}$ and the iterative one by I-FGSM$_{\varepsilon}^{\#}$, $\# \in \{ \_,ll \}$, where the superscript discriminates between untargeted and targeted (here $ll$ refers to "least likely"). 
For the re-implementation of SSMM and DNNM \cite{Metzen2017}, we use the parameters $\varepsilon = 0.1\cdot255$, $\alpha = 0.01\cdot255$, $n = 60$ and $\tau = 0.75$. For SSMM, the target image is chosen randomly for both datasets and for DNNM, the class person is to be deleted for the Cityscapes dataset. 
For the VOC dataset, the DNNM attack makes no sense, since on the input images often only one object or several objects of the same class are contained.
For our experiments, we use a model zoo\footnote{\url{https://github.com/LikeLy-Journey/SegmenTron}} where we add the implementations of the adversarial attacks FGSM, I-FGSM, SSMM and DNNM. As we also use the pre-trained networks provided in the repository, we run experiments for the Cityscapes dataset on both models, DeepLabv3+ and HRNet, and for VOC on the DeepLabv3+ network.

For the patch attack introduced in \cite{Nesti2022}, we use the provided code with default parameters and consider two different segmentation models, BiSeNet and DDRNet, applied to the one tested real world dataset (Cityscapes).
Since we use the cross-attack procedure (logistic regression) as detection model, i.e., we train the classifier on clean and perturbed data attacked by an attack other than the patch, we use the data obtained from the DeepLabv3+ to train the detector and test on the DDRNet. For the HRNet and the BiSeNet we proceed analogously, since in each case the prediction performance (in terms of mIoU) of both networks is similar.

As the Cityscapes dataset provides high-resolution images ($1024 \times 2048$ pixels) which require a great amount of memory to run a full backward pass for the computation of adversarial samples, we re-scale image size for this dataset to $512 \times 1024$ when evaluating. In Figure~\ref{fig:preds_adv}, a selection of these attacks is shown for the Cityscapes dataset and the DeepLabv3+ network (or DDRNet for the patch attack).
%
%
\paragraph{Evaluation Metrics}
Our detection models provide per image a probability $p(x)$ of being clean (and not perturbed). 
The image is then classified as attacked if the probability exceeds a threshold $\kappa$ for which we tested $40$ different values equally spaced in $ [0,1]$.
The first evaluation metric we use is the averaged detection accuracy (ADA) which is defined as the proportion of images that are classified correctly. 
As this metric depends on a threshold $\kappa$, we report the optimal ADA score obtained by $\ada = \max_{\kappa \in [0,1]} \text{ADA}(\kappa)$.
Secondly, we compute the area under the receiver operating characteristic curve ($\auc$) to obtain a threshold independent metric.
Lastly, we consider the true positive rate while fixing the false positive rate on clean images to $5\%$ ($\tpr$).
%
%
\subsection{Numerical Results}\label{sec:exp_results}
In the following, we study the attack success performance and evaluate our adversarial attack detection performance.

\begin{table}
\caption{APSR results for the semantic segmentation predictions on clean data.}
\centering
\scalebox{0.8}{
\begin{tabular}{c|cccc}
\cline{1-5}
\multicolumn{1}{c}{} & DeepLabv3+ & DDRNet & HRNet & BiSeNet \\
\cline{1-5}
Cityscapes & $6.84$ & $4.00$ & $5.48$ & $5.26$  \\ 
VOC & $2.92$ & - & - & -  \\
\cline{1-5}
\end{tabular} }
\label{tab:apsr_nn}
\end{table} 
\begin{figure}
    \centering
    \includegraphics[trim=0 10 0 5,clip,width=0.47\textwidth]{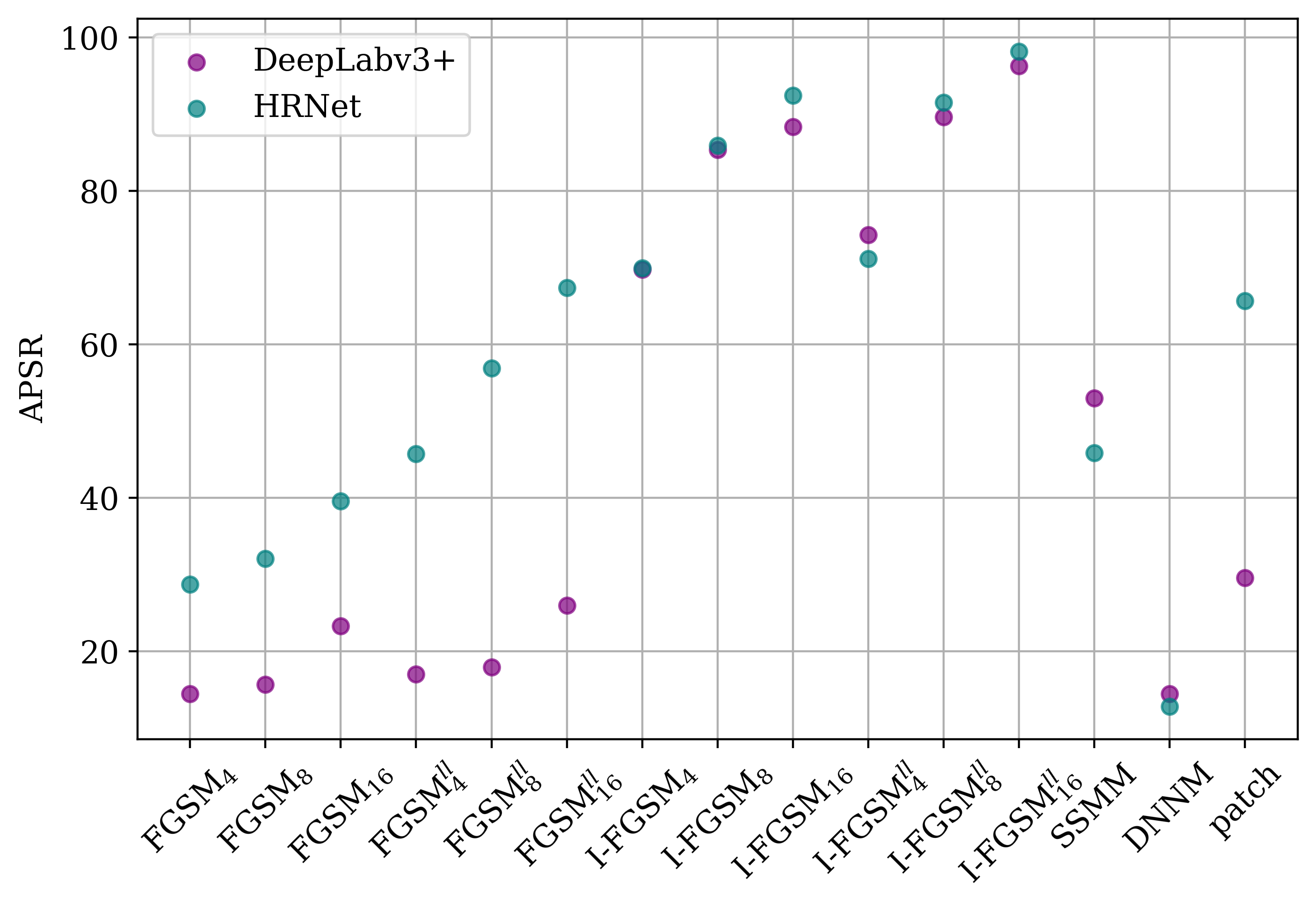} 
    \caption{APSR results for the Cityscapes dataset and both networks perturbed by different attacks.}
    \label{fig:apsr_cs}
\end{figure}
%
%
\paragraph{Attack Performance}
In order to access the performance, i.e., the strength, of the attack generating methods, we consider the attack pixel success rate (APSR) \cite{Rony2022} defined by
\begin{equation}\label{eq:apsr}
    \text{APSR} = \frac{1}{|\mathcal{I}|} \sum_{(i,j) \in \mathcal{I}} \argmax_{y\in\mathcal{C}} f(x^{adv};w)_{ij}^{y} \neq y_{ij}^{GT} \enspace ,
\end{equation}
where $y_{ij}^{GT}$ denotes the true class of the pixel at location $(i,j)$.
Note, this metric is the opposite of the accuracy, as it focuses on falsely (and not correct) predicted pixels.
If we replace $x^{adv}$ in eq.~(\ref{eq:apsr}) by the input image $x$, we obtain a measure how well the semantic segmentation model performs on clean data. 
The values for this measure for the different networks and both datasets are given in Table~\ref{tab:apsr_nn}. As expected, we observe small APSR scores: all values are below $6.84\%$.

\begin{table}[t]
\caption{APSR results for the VOC dataset and the DeepLabv3+ network perturbed by different attacks.} 
\centering
\scalebox{0.8}{
\begin{tabular}{ccccc}
\cline{1-5}
FGSM$_{4}$ & FGSM$_{4}^{ll}$ & I-FGSM$_{4}$ & I-FGSM$_{4}^{ll}$ & SSMM \rule{0mm}{2.5mm}\\
$17.81$ & $19.68$ & $56.73$ & $64.30$ & $62.99$ \\
\cline{1-5}
FGSM$_{8}$ & FGSM$_{8}^{ll}$ & I-FGSM$_{8}$ & I-FGSM$_{8}^{ll}$ &  \rule{0mm}{2.5mm}\\
$17.93$ & $19.66$ & $74.94$ & $82.51$ &  \\
\cline{1-5}
FGSM$_{16}$ & FGSM$_{16}^{ll}$ & I-FGSM$_{16}$ & I-FGSM$_{16}^{ll}$ & \rule{0mm}{2.5mm}\\
$16.67$ & $17.65$  & $80.98$ &  $91.68$ & \\
\cline{1-5}
\end{tabular} }
\label{tab:apsr_voc}
\end{table}

The APSR results for various attacks on the Cityscapes dataset are shown in Figure~\ref{fig:apsr_cs}.
For all variations of the FGSM attack (untargeted vs.\ targeted, non-iterative vs.\ iterative) the APSR increases with larger magnitude of perturbation strength. 
Moreover, targeted attacks lead to larger ASPR values than their untargeted counterpart.
The I-FGSM outperforms the FGSM due to the iterative procedure of individual perturbation steps.
For the SSMM attack, the target is a randomly chosen image from the dataset. 
The examples shown in Figure~\ref{fig:preds_adv} (a) and (f) indicate, that the correct and target classes 
of the clean and the perturbed image coincide in several areas, such as the street or the sky reflecting the nature of street scenes. 
This may explain the relatively low ASPR values around $50\%$.
For the DNNM attack, the APSR scores are comparatively small since most parts of the images are not perturbed but only one class is to be deleted. 
We observe that the performance of the patch attack has more or less impact depending on the model.
Comparing the two models, we find that DeepLabv3+ is more robust against adversarial attacks, as the APSR values are mostly smaller than those of the HRNet network. 
A selection of qualitative results is shown in Appendix~\ref{sec:app_examples}.

The results for the VOC dataset  given in Table~\ref{tab:apsr_voc} are qualitatively similar to the findings for the Cityscapes dataset.
However, the outcome for the (targeted as well as untargeted) non-iterative FGSM attack differs, i.e., the APSR scores are not increasing with higher noise but stay at similar values. 
This observation is confirmed in the sample images in the Appendix~\ref{sec:app_examples} which show very little variation over the various magnitudes of noise. 
In summary, for both datasets and the different network architectures, most attacks achieve high APSR values and greatly alter the prediction. Thus, the detection of such attacks is extremely important.
%
%
\paragraph{Evaluation of our Detection Method}
\begin{figure}
    \centering
    \includegraphics[trim=0 10 0 5,clip,width=0.47\textwidth]{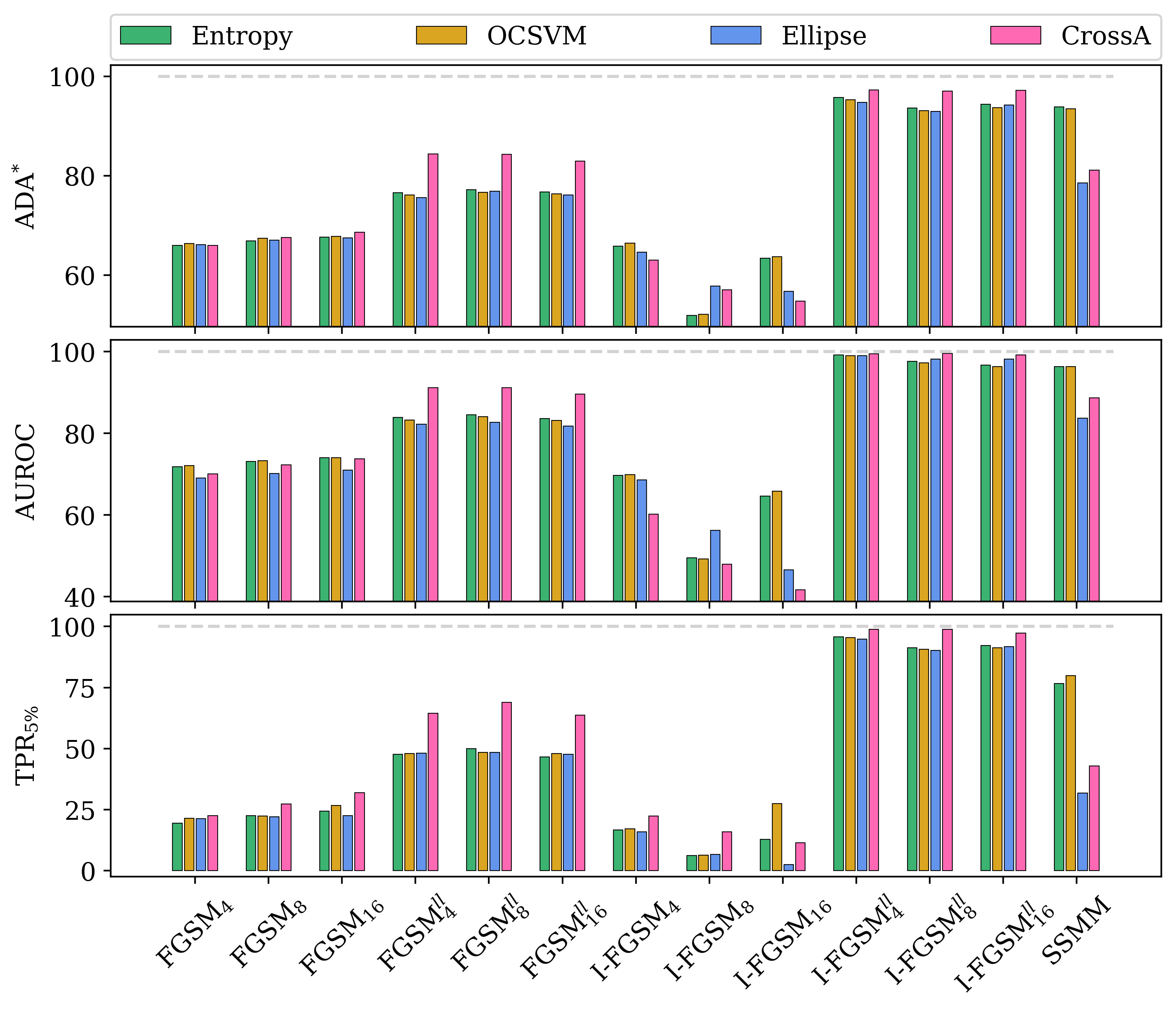} 
    \caption{ Detection performance results for the VOC dataset and the DeepLabv3+ network.}
    \label{fig:metrics_voc}
\end{figure}
\begin{figure*}
    \centering
    \includegraphics[trim=0 10 0 5,clip,width=0.47\textwidth]{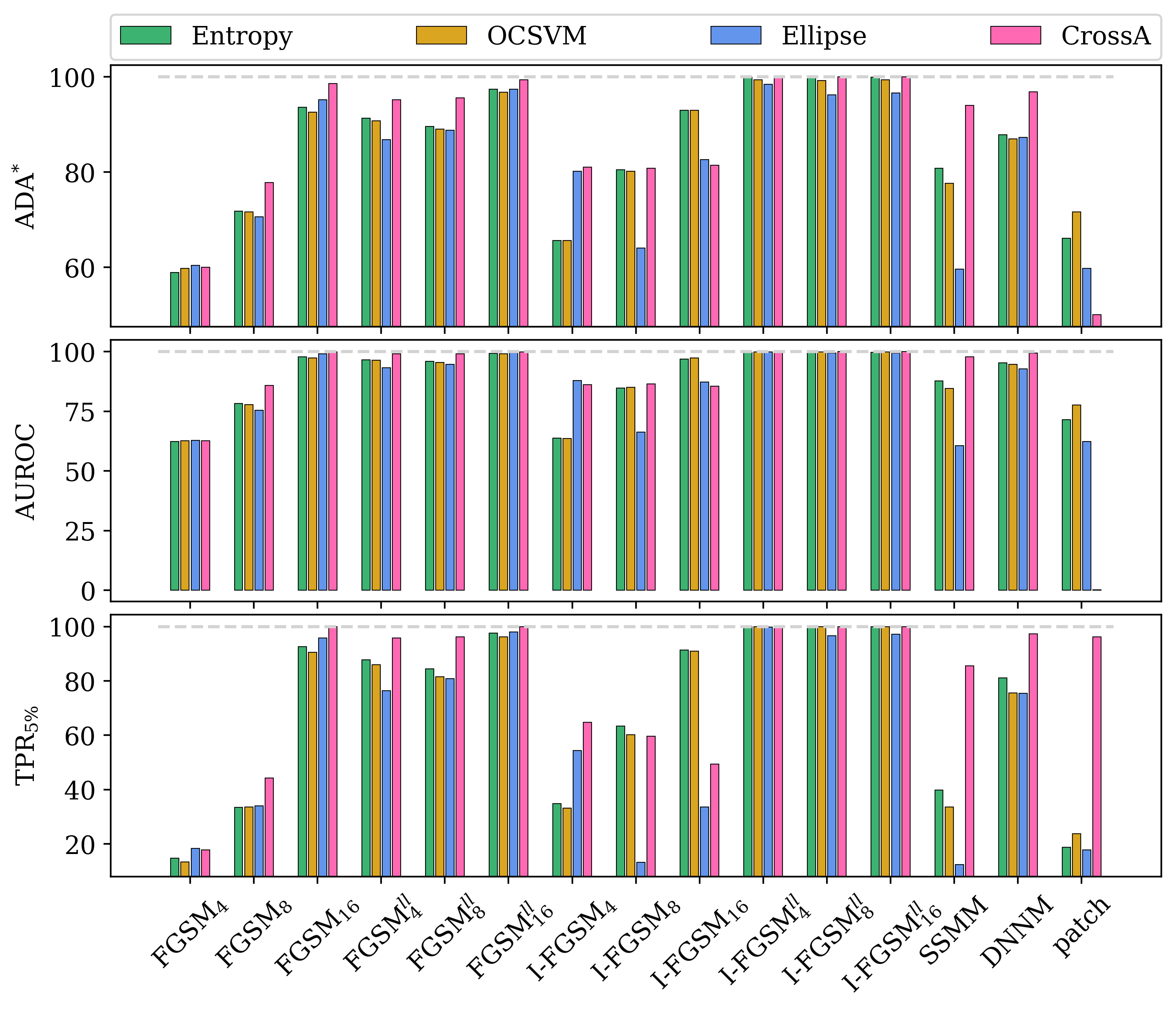} 
    \includegraphics[trim=0 10 0 5,clip,width=0.47\textwidth]{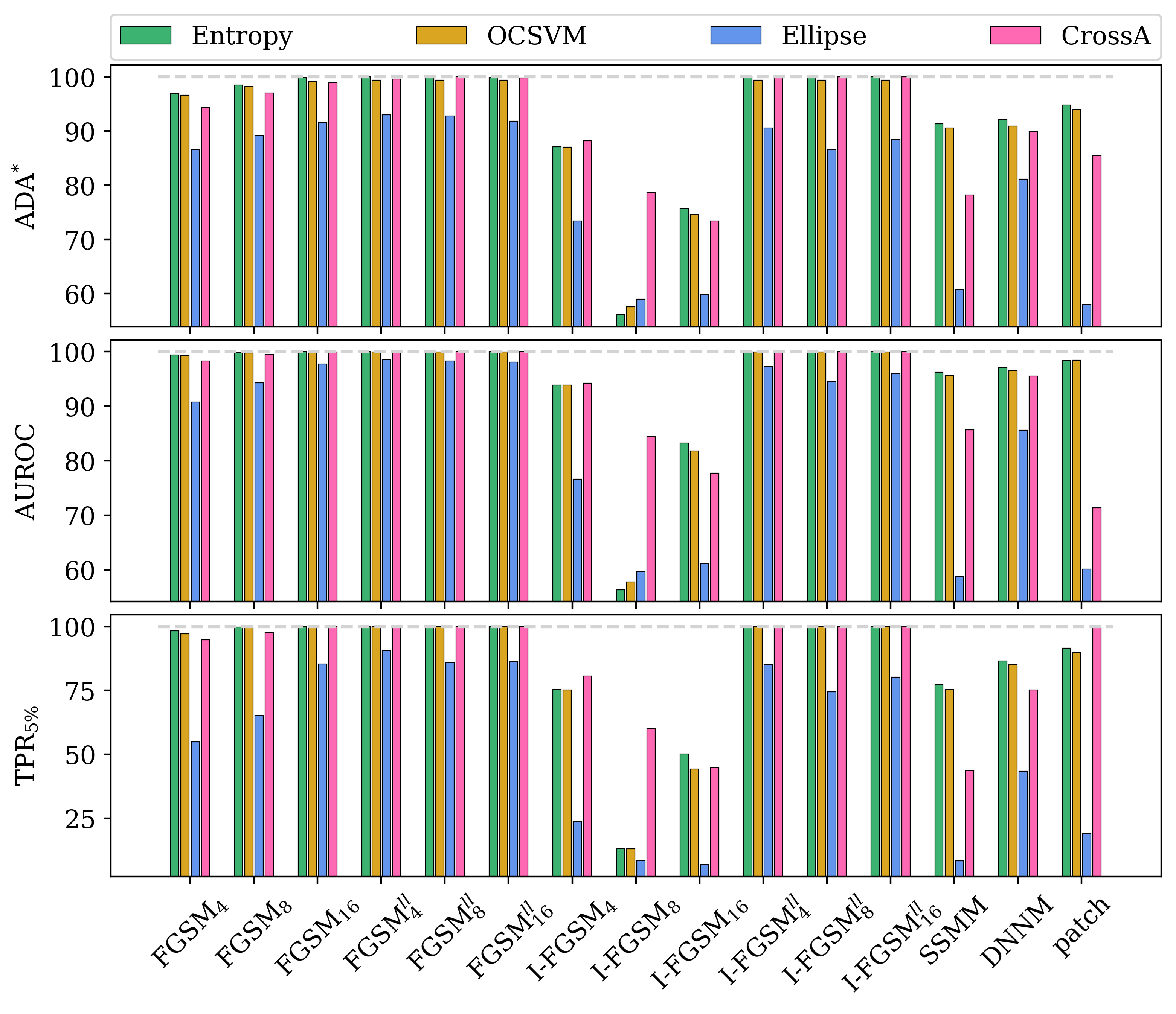} 
    \caption{ Detection performance results for DeepLabv3+ (\emph{left}) and the HRNet (\emph{right}) trained on  the Cityscapes dataset.}
    \label{fig:metrics_cs}
\end{figure*}
The defense approaches described 
above are created for and tested only on a specific type of attacks. 
The sole presented detection approach \cite{Xiao2018} is only tested on stationary segmentation mask methods and is computationally expensive due to the requirement of multiple runs of the network. 
For this reason, neither this detection approach nor the defense methods can be considered as suitable baselines.  

In the following, we denote the single feature, mean entropy based classification by \emph{Entropy}, the ordinary one-class support vector machine by \emph{OCSVM}, the outlier detection method proposed in \cite{Rousseeuw1999}
by \emph{Ellipse} and the logistic regression model by \emph{CrossA}. For training the regression model, we chose data resulting from the iterative targeted FGSM attack with a magnitude of noise of $2$ as perturbed data (assuming that it might be advantageous to have malicious data stemming from an attack method with small perturbation strength, which makes the attack harder to detect). 
The resulting classifier is then evaluated against all other attacks. 
Note, the training of the detection models is light-weight and no knowledge of the process for generating adversarial examples is needed. For evaluating our detection models, we use cross-validation with $5$ runs.

The detection results for the VOC dataset are shown in Figure~\ref{fig:metrics_voc} and for the Cityscapes dataset in Figure~\ref{fig:metrics_cs}.
We observe comparatively lower detection performance results for smaller perturbation magnitudes for the untargeted 
FGSM attacks which may be explained by the fact that weaker attacks lead to a change of prediction for a lower number of pixels and are thus more difficult to detect. 
Targeted attacks are better detected than untargeted attacks (with ADA$^*$ values over $80\%$ for all models). This could be due to the procedure of picking the most unlikely class as target which results in larger changes of the uncertainty measures used as features. 
The detectors perform not that well on the adversarial examples resulting from the untargeted I-FGSM despite the strength of the attack. An inspection of these examples shows that during segmentation only a few classes are predicted (see Figure~\ref{fig:preds_adv} (c)) with often low uncertainty for large connected components which complicates the distinction between clean and perturbed data. 
Interesting are the high detection results of up to $96.89\%$ $\ada$ values (obtained by CrossA for the Cityscapes dataset and the DeepLabv3+ network) for the DNNM attack as the perturbation targets only a few pixels, (APSR values around $15\%$) and is therefore difficult to detect. 
For the patch attack, it is noticeable that the detection performance for the DeepLabv3+ network on the Cityscapes dataset is low compared to the HRNet which is explained by the higher disturbance power of this attack on the HRNet, reflected in $36.11$ percentage points higher APSR values. 
In general, the detection capability for attacks on the HRNet is stronger than for the DeepLabv3+, since the HRNet is more easily to attack (see higher APSR  values in Figure~\ref{fig:apsr_cs}). 
Generally, our experiments highlight the high potential of investigating 
uncertainty information for successfully detecting adversarial segmentation attacks. Already the basic method \emph{Entropy}
leads to high accuracies often outperforming OCSVM and Ellipse. However, across different attacks and datasets, CrossA achieves $\ada$ values of up to $100\%$. Thus, our light-weight and uncertainty-based detection approach should be considered as baseline for future methods.
%
%
%
\section{\uppercase{Conclusion}}\label{sec:conc}
In this work, we introduced a new uncertainty-based approach for the detection of adversarial attacks on semantic image segmentation tasks. 
We observed that uncertainty information as given by the entropy behaves differently on clean and perturbed images and used this property to distinguish between the two cases with very basic classification models. Our approach works in a light-weight and post-processing manner, i.e., we do not modify the model nor need knowledge of the process used by the attacker for generating adversarial examples. We achieve averaged detection accuracy values of up to $100\%$ for different network architectures and datasets.
Moreover, it has to be pointed out, that our proposed detection approach is the first that was not designed for a specific adversarial attack, but has a high detection capability across multiple types. Given the high detection accuracy and the simplicity of the proposed approach, we are convinced, that it should serve as simple baseline for more elaborated but computationally more expensive approaches developed in future.
%
\section*{ACKNOWLEDGEMENTS} 
This work is supported by the Ministry of Culture and Science of the German state of North Rhine-Westphalia as part of the KI-Starter research funding program and by the Deutsche Forschungsgemeinschaft (DFG, German Research Foundation) under Germany’s Excellence Strategy – EXC-2092 CASA – 390781972.

\bibliographystyle{apalike}
{\small
\bibliography{example}}

\newpage
\appendix
\section*{\uppercase{APPENDIX}}
\section{More Adversarial Examples}\label{sec:app_examples}
\begin{figure*}
    \centering
    \includegraphics[width=0.92\textwidth]{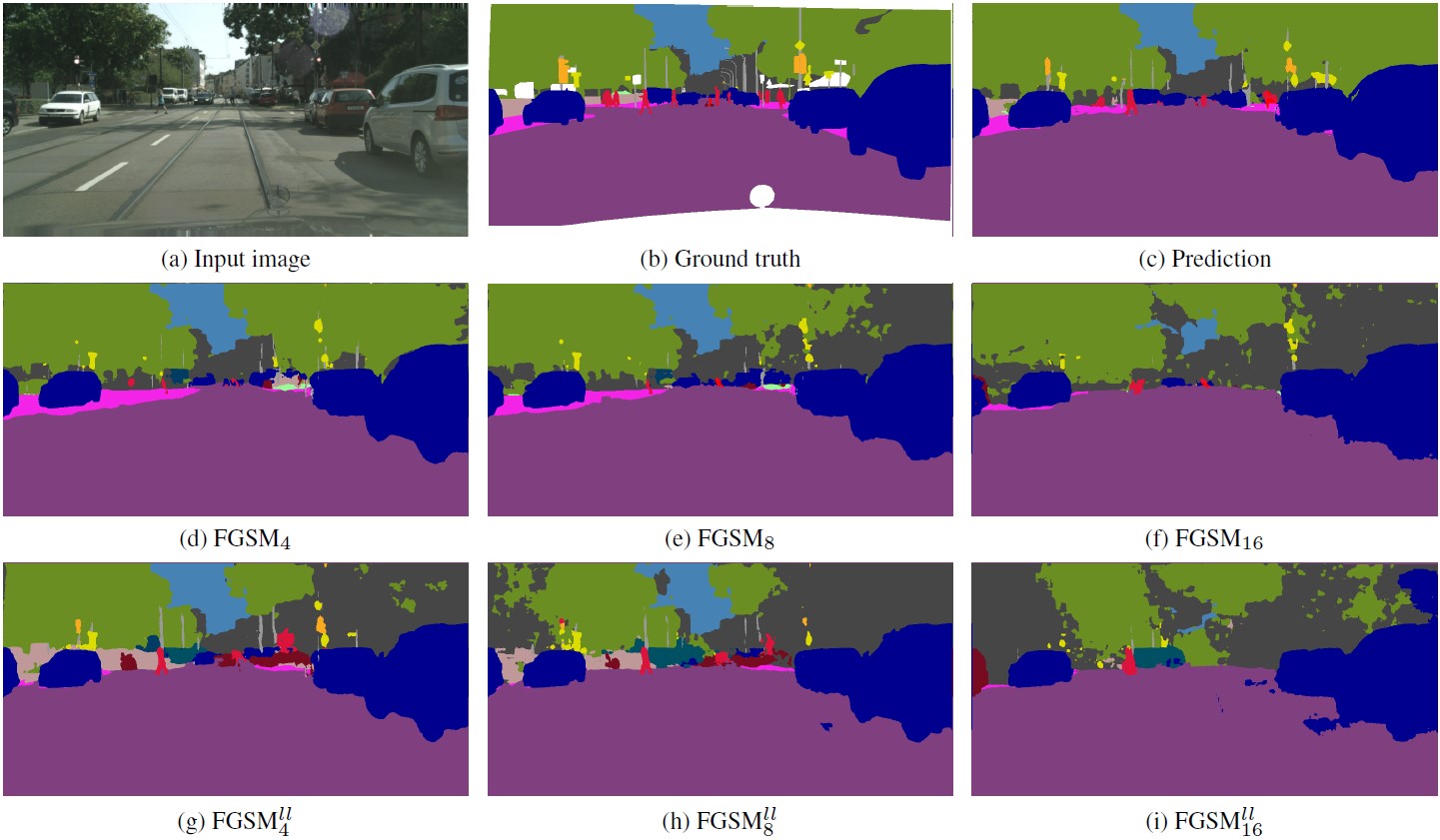}
    \includegraphics[width=0.92\textwidth]{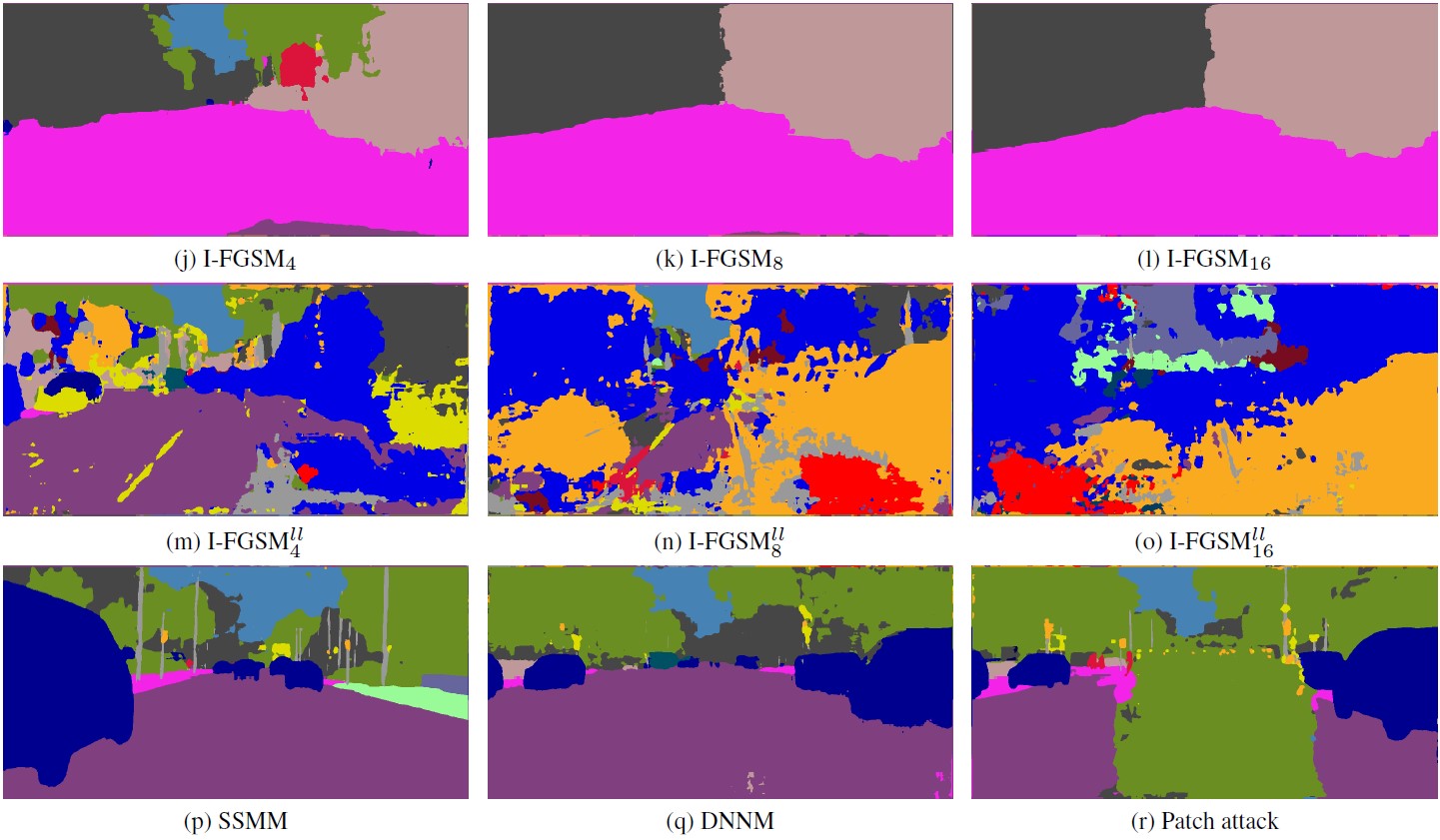}
    \caption{Input image (a) with corresponding ground truth (b) for the Cityscapes dataset. Semantic segmentation prediction obtained by the DeepLabv3+ network for a clean image (c) and perturbed images generated by various attacks (d)-(r).}
    \label{fig:app_adv_cs_dl}
\end{figure*}

\begin{figure*}
    \centering
    \includegraphics[width=0.92\textwidth]{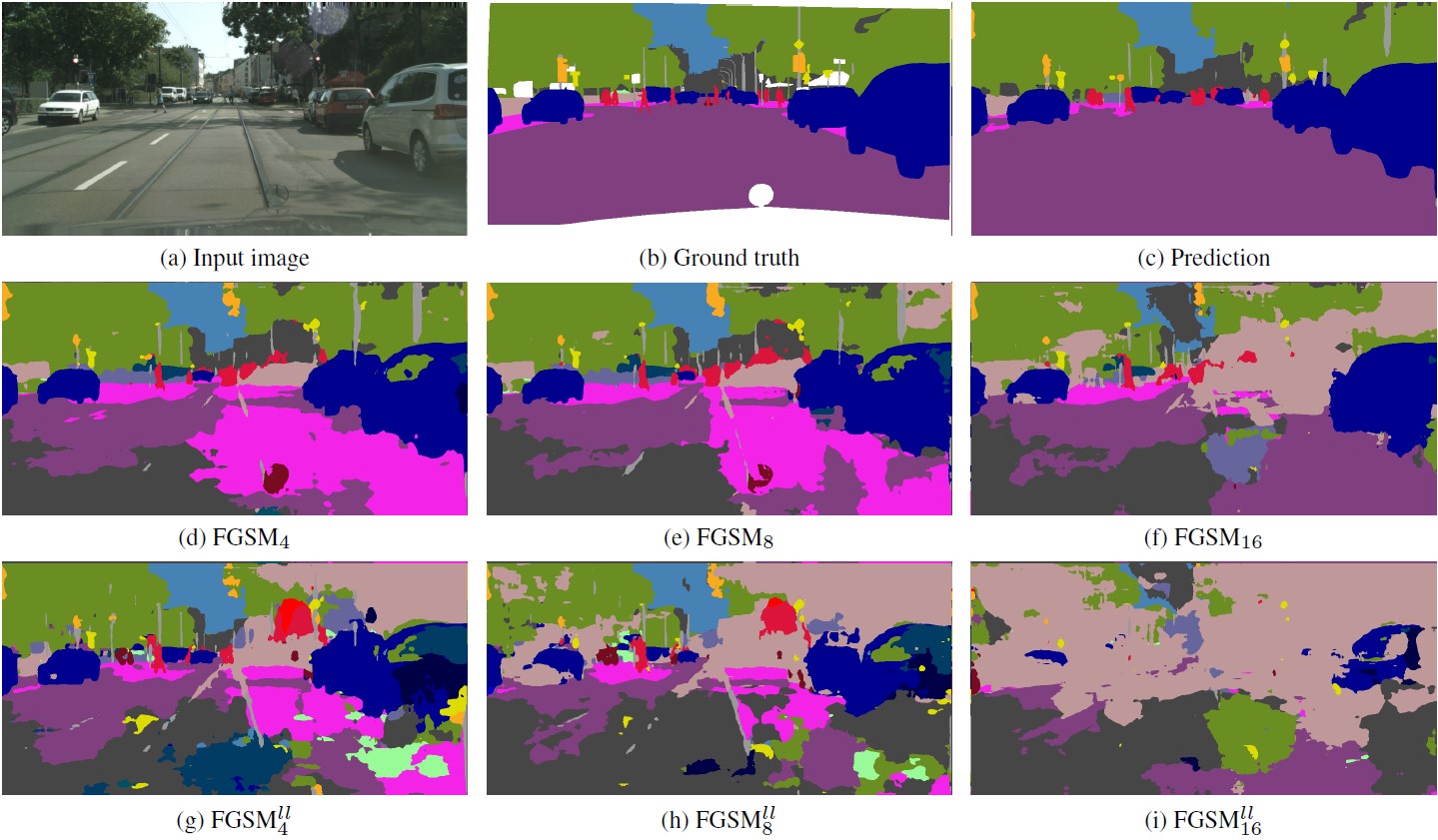}
    \includegraphics[width=0.92\textwidth]{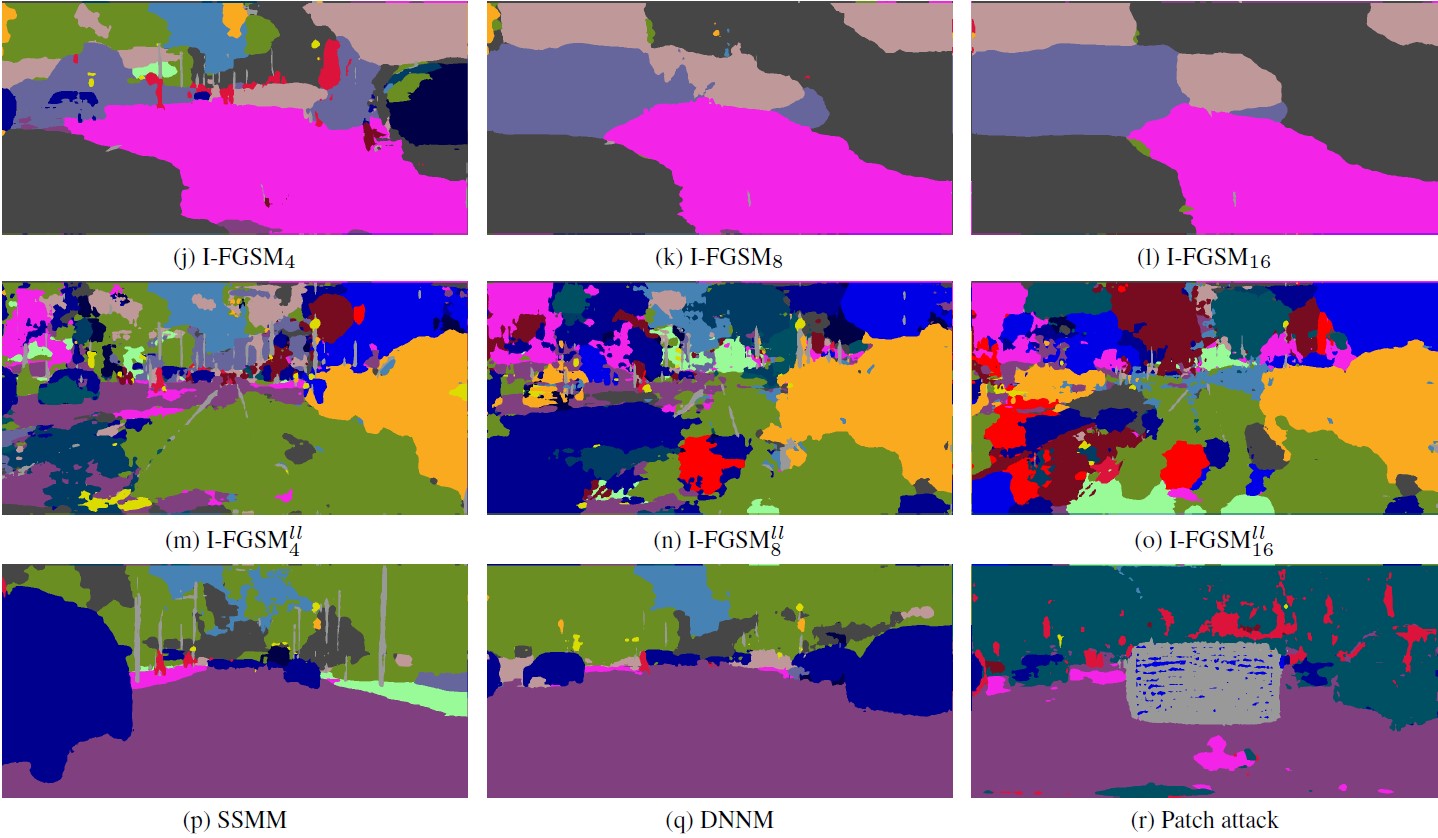}
    \caption{Input image (a) with corresponding ground truth (b) for the Cityscapes dataset. Semantic segmentation prediction obtained by the HRNet network for a clean image (c) and perturbed images generated by various attacks (d)-(r).}
    \label{fig:app_adv_cs_hr}
\end{figure*}

\begin{figure*}
    \centering
    \includegraphics[width=0.92\textwidth]{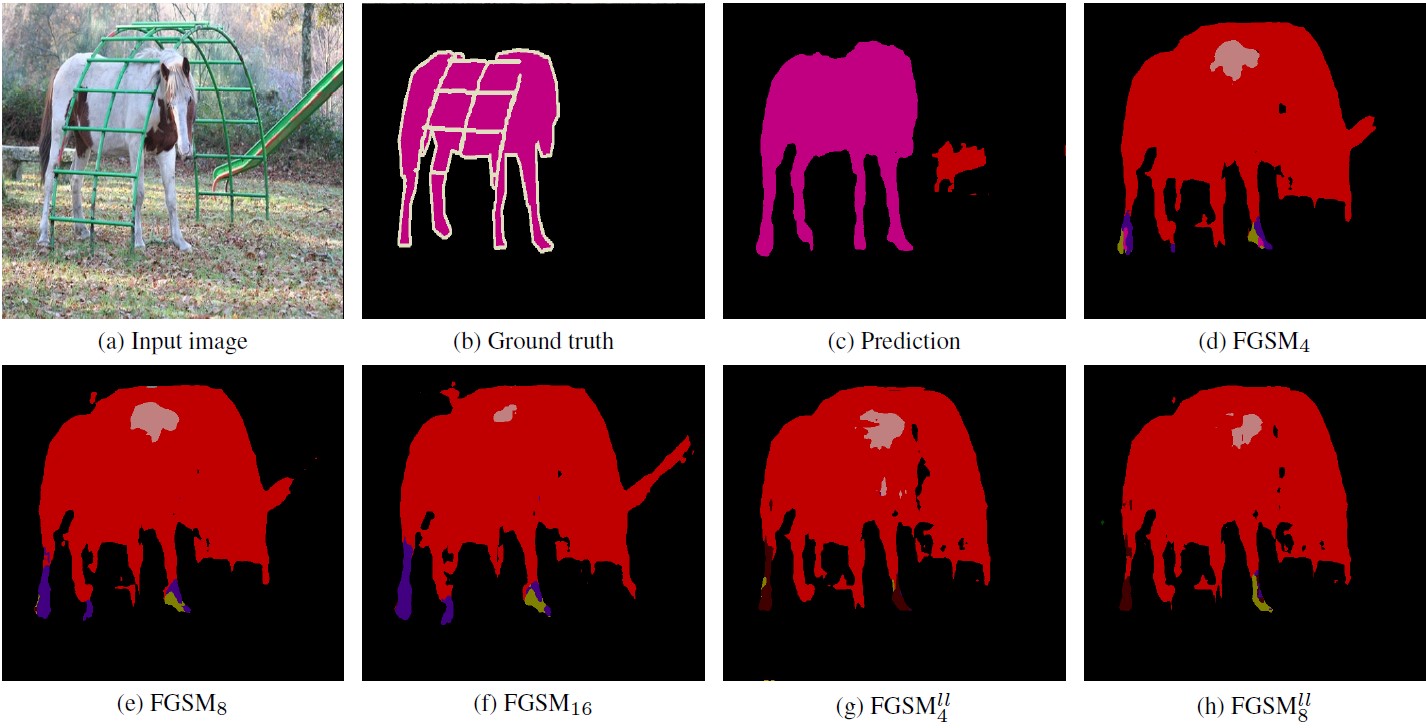}
    \includegraphics[width=0.92\textwidth]{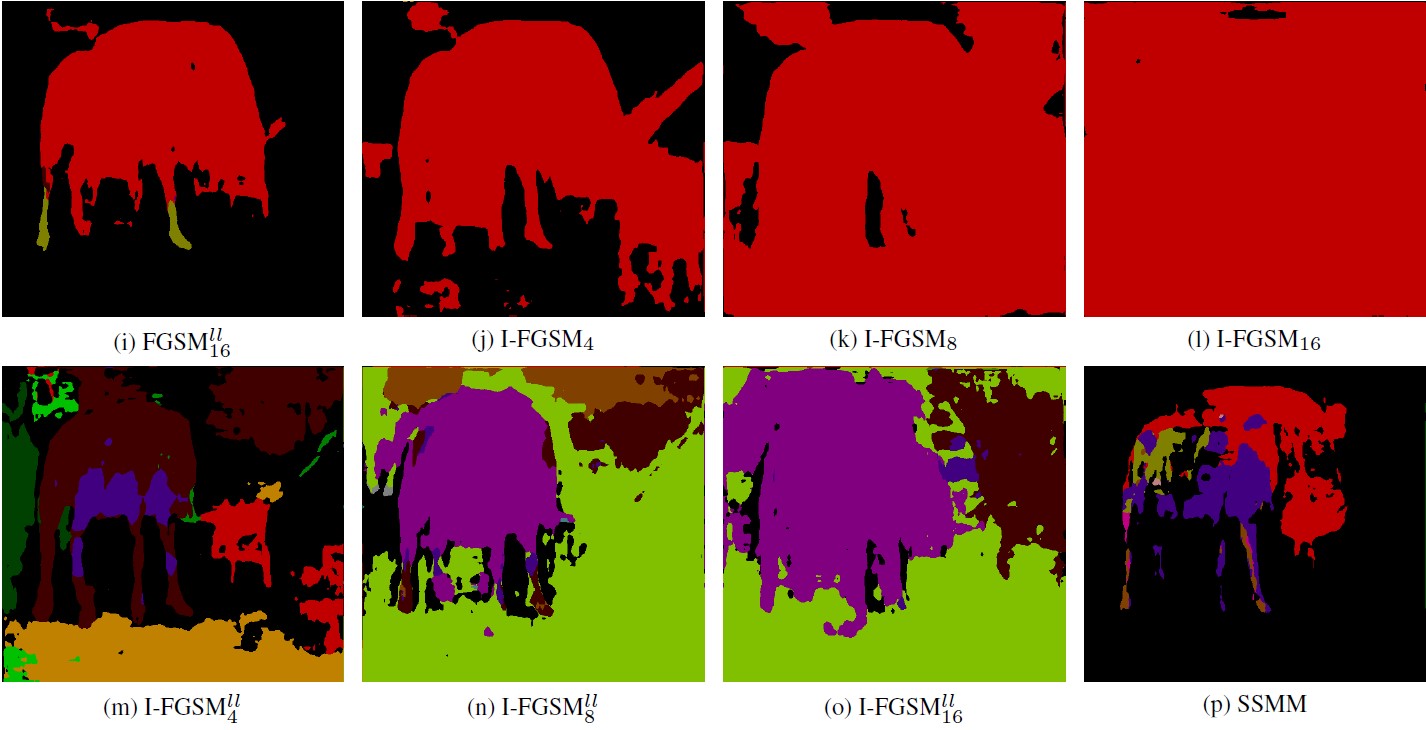}
    \caption{Input image (a) with corresponding ground truth (b) for the VOC dataset. Semantic segmentation prediction obtained by the DeepLabv3+ network for a clean image (c) and perturbed images generated by various attacks (d)-(p).}
    \label{fig:app_adv_voc_dl}
\end{figure*}

In this section, we provide qualitative results for the considered attacks. In \autoref{fig:app_adv_cs_dl} and \autoref{fig:app_adv_cs_hr} semantic segmentation predictions for an example image from the Cityscapes dataset are shown and in \autoref{fig:app_adv_voc_dl} for an example image from the VOC dataset.
For Cityscapes, we observe more pixel changes for the FGSM attack for increasing noise. For the untargeted I-FGSM attack, the predictions result in less different classes across datasets and network architectures. In general, for the weaker HRNet in comparison to the DeepLabv3+, the perturbations are more visible. Even the non-iterative FGSM attacks and also the patch attack show great success. 

\begin{figure*}
    \centering
    \includegraphics[width=0.92\textwidth]{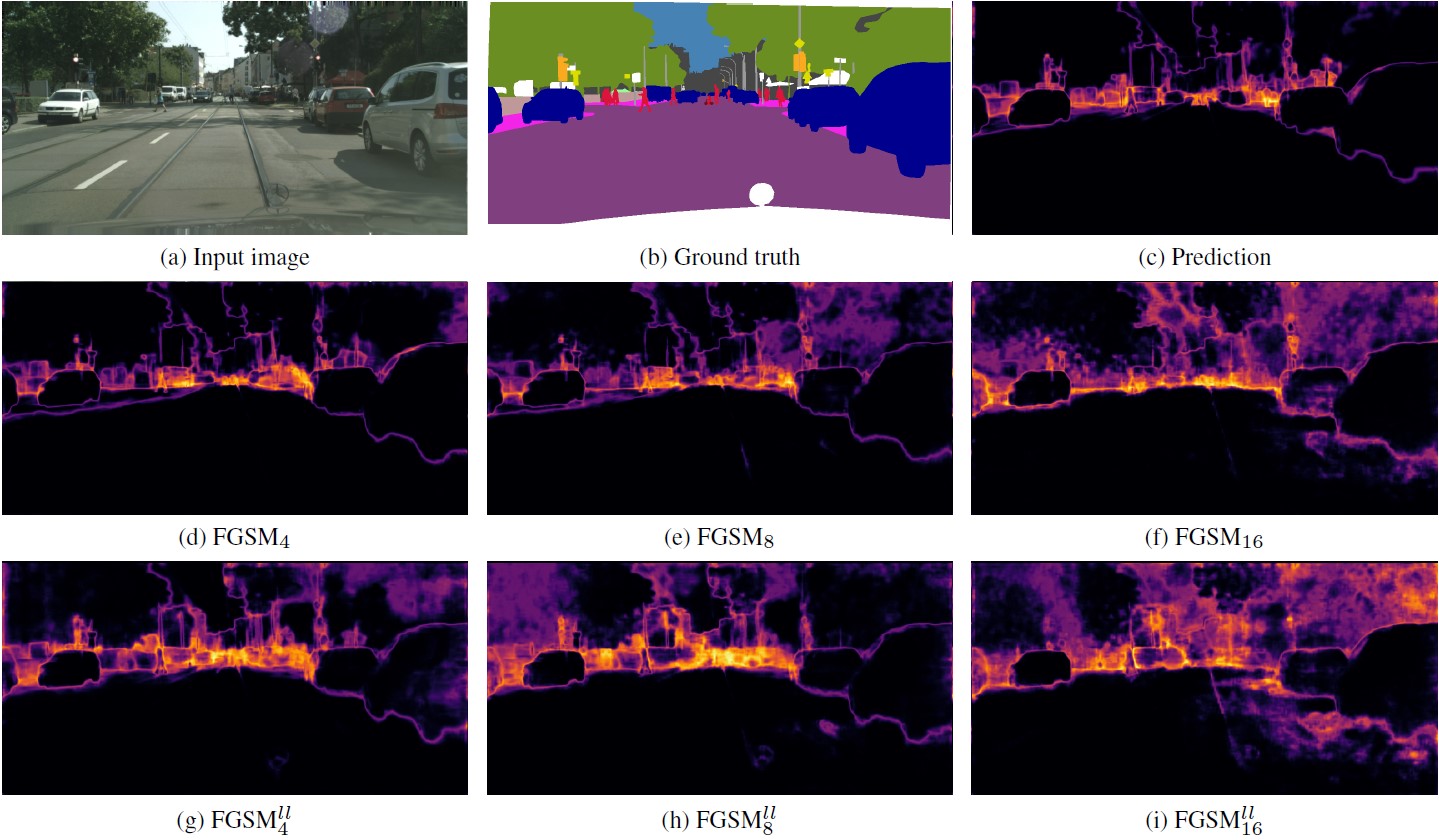}
    \includegraphics[width=0.92\textwidth]{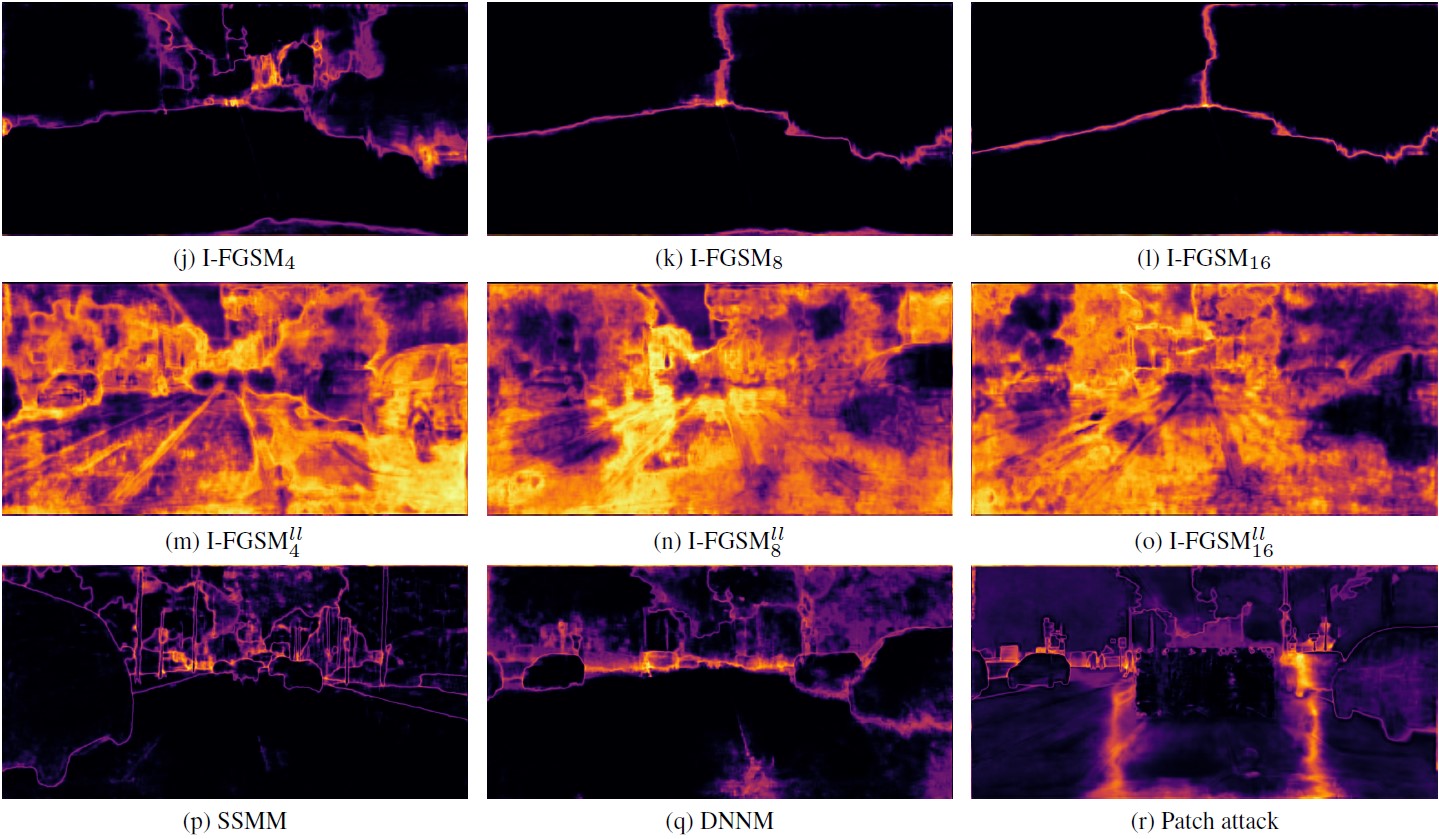}
    \caption{Input image (a) with corresponding ground truth (b) for the Cityscapes dataset. Entropy heatmaps obtained by the DeepLabv3+ network for a clean image (c) and perturbed images generated by various attacks (d)-(r).}
    \label{fig:app_hm_cs_dl}
\end{figure*}

\begin{figure*}
    \centering
    \includegraphics[width=0.92\textwidth]{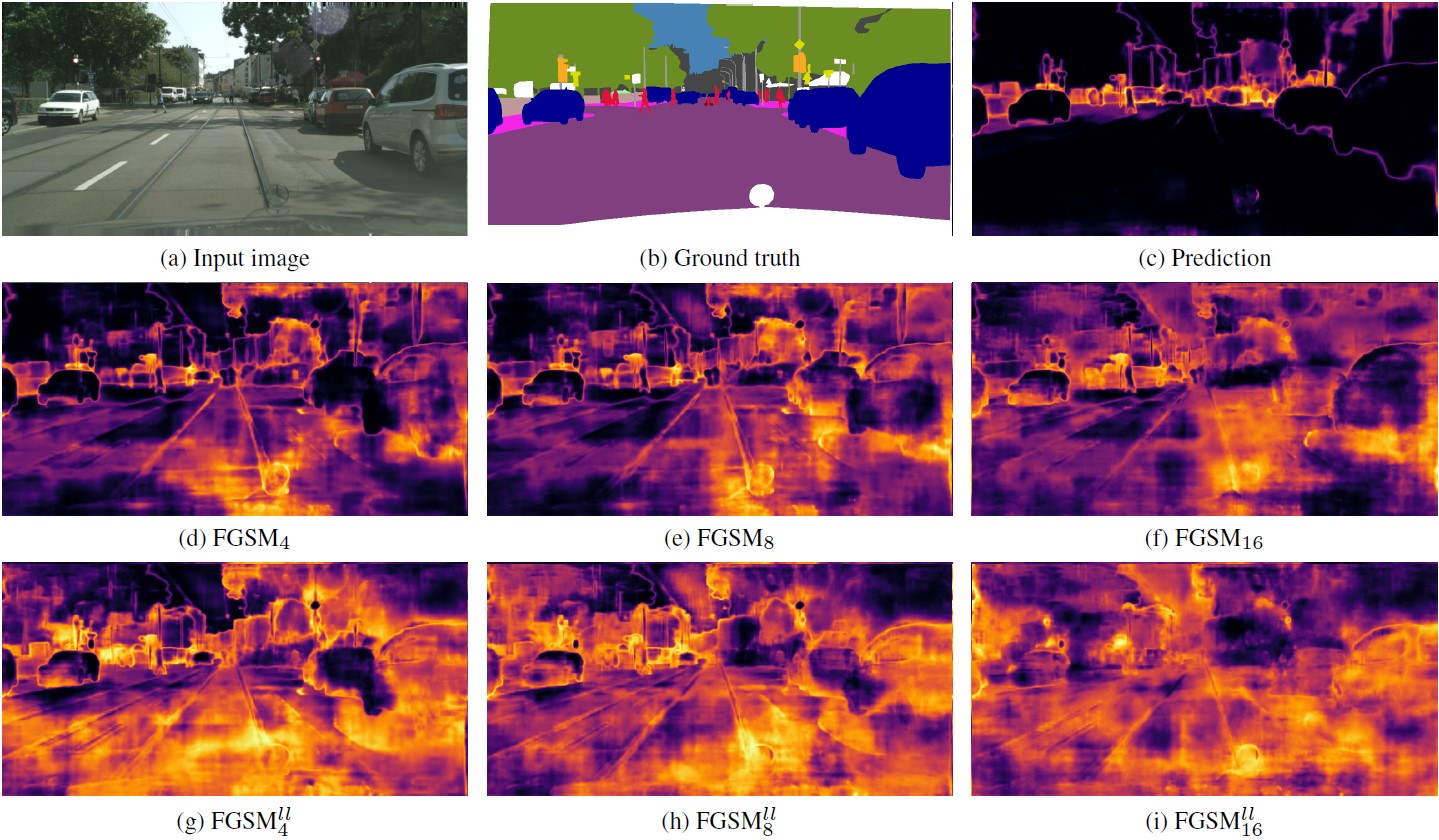}
    \includegraphics[width=0.92\textwidth]{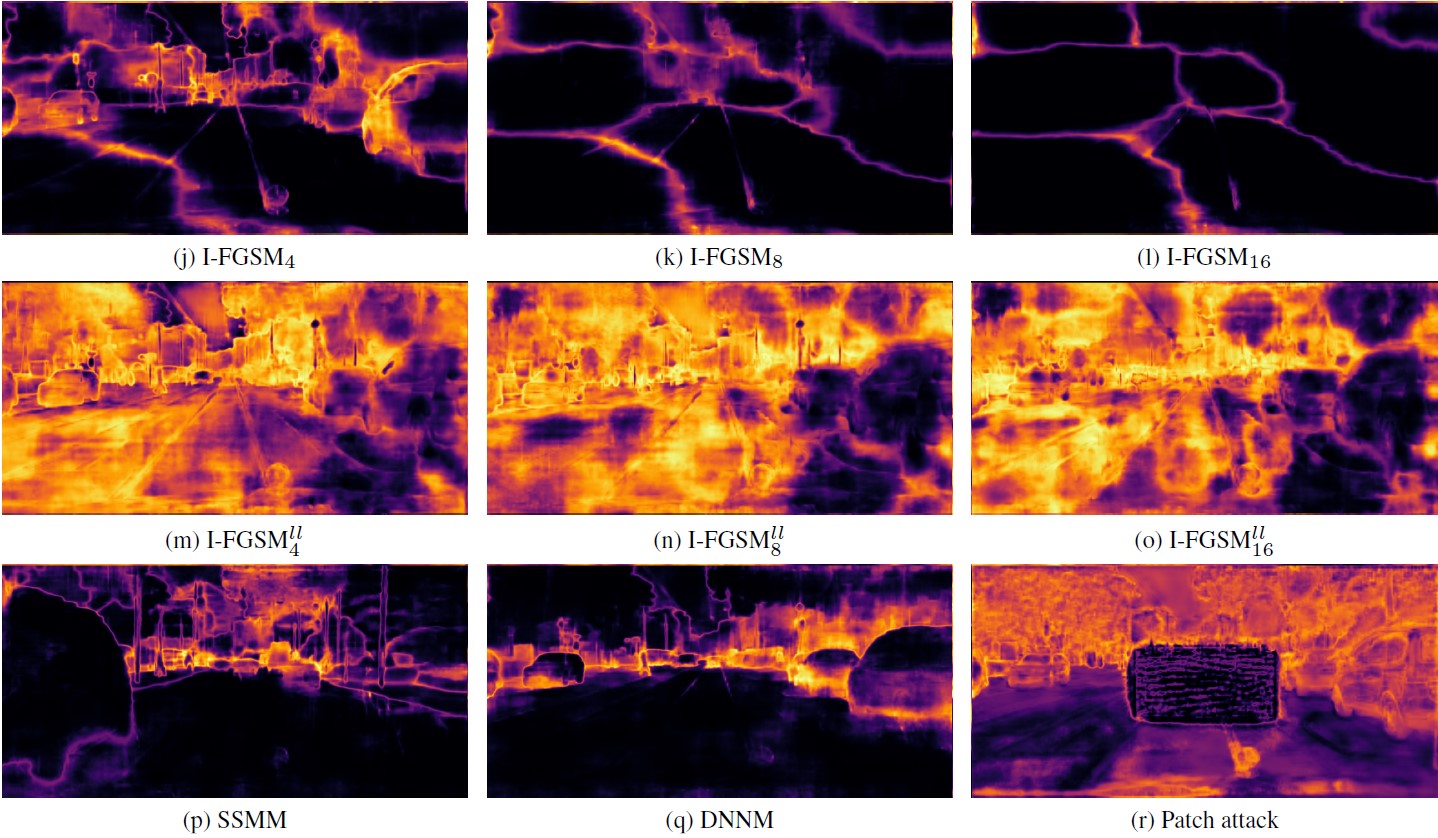}
    \caption{Input image (a) with corresponding ground truth (b) for the Cityscapes dataset. Entropy heatmaps obtained by the HRNet network for a clean image (c) and perturbed images generated by various attacks (d)-(r).}
    \label{fig:app_hm_cs_hr}
\end{figure*}

\begin{figure*}
    \centering
    \includegraphics[width=0.92\textwidth]{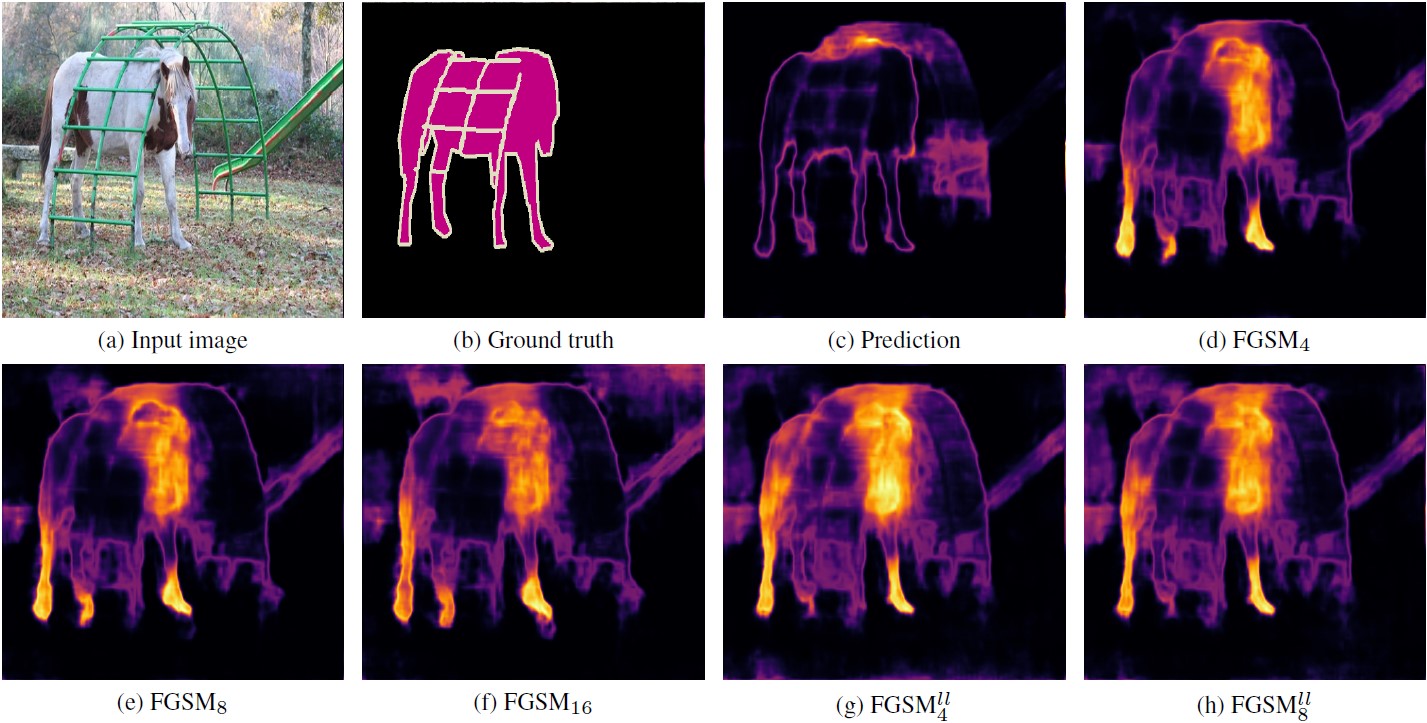}
    \includegraphics[width=0.92\textwidth]{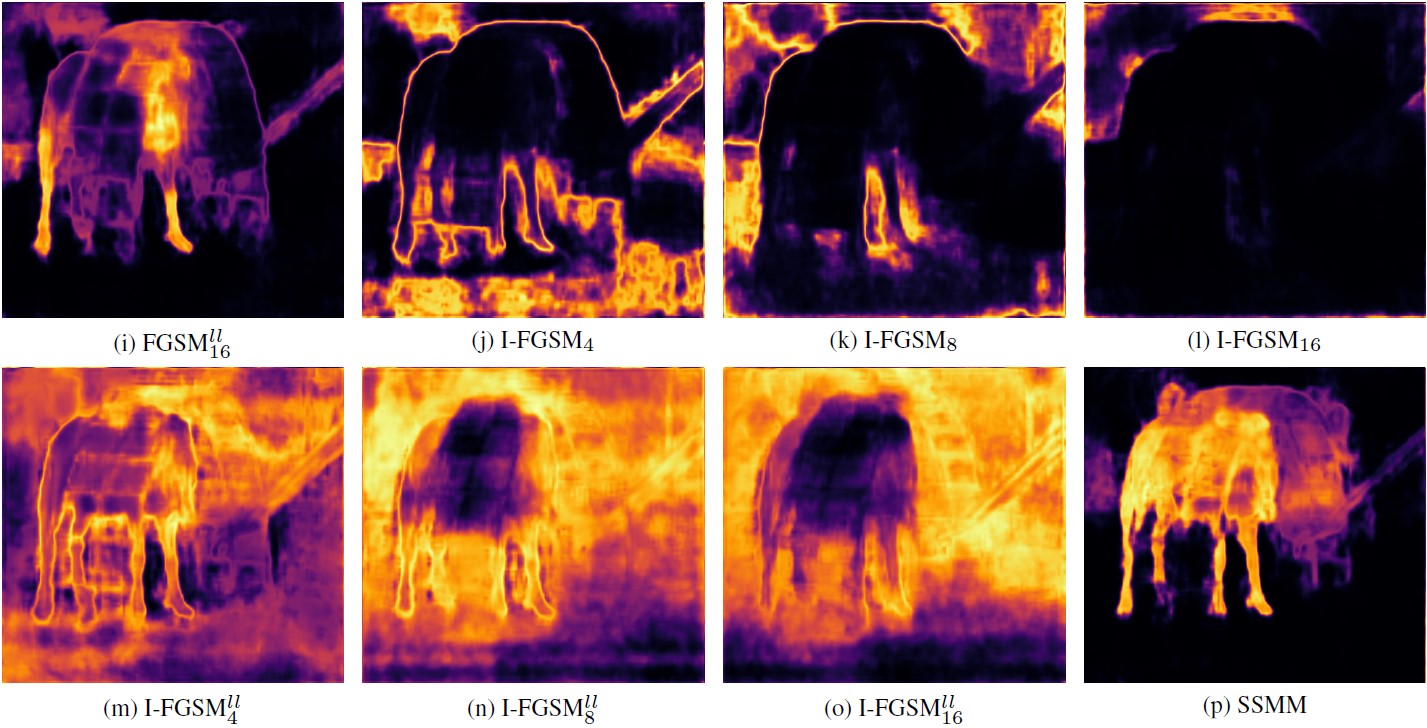}
    \caption{Input image (a) with corresponding ground truth (b) for the VOC dataset. Entropy heatmaps obtained by the DeepLabv3+ network for a clean image (c) and perturbed images generated by various attacks (d)-(p).}
    \label{fig:app_hm_voc_dl}
\end{figure*}

In \autoref{fig:app_hm_cs_dl}, \autoref{fig:app_hm_cs_hr}, and \autoref{fig:app_hm_voc_dl}, the corresponding entropy heatmaps for the clean images and the perturbed ones are given. 
The heatmaps for the targeted and untargeted FGSM attack still look similar to the heatmap for the clean image. These attacks change the prediction though, however, shapes are still recognizable. High uncertainties can also be seen in the areas where the attack was successful. 
For the iterative untargeted FGSM attack, the heatmaps are dark with only high values on the segment boundaries. In this case, the perturbation can be detected as the average uncertainty per image deviates downward from the clean prediction. 
The iterative targeted attack shows the highest uncertainty, since the predictions consist of comparatively large numbers of different segments and classes.
Similar to the clean image, the SSMM and DNNM attacks indicate higher uncertainties in the background while the certainty in the prediction of the road and the cars is high.
The patch attack has different effects on both networks, i.e. for the DeepLabv3+, is uncertainty higher in general and especially at the patch boundaries and for the HRNet, the impact on the prediction is stronger, which is also reflected in the heatmaps.

\end{document}